\documentclass[journal]{IEEEtran} 
\usepackage{cite}

\usepackage{changes}
\usepackage{graphicx}
\usepackage{multirow}

\ifCLASSINFOpdf
\else
\fi
\usepackage{amssymb,amsmath}
\usepackage{algorithmic}
\usepackage[ruled,vlined]{algorithm2e}
\DeclareMathOperator*{\argmax}{arg\,max}

\begin{document}
%
\title{Efficient Multi-Objective Optimization through Population-based Parallel Surrogate Search}
%
%
%

\author{Taimoor~Akhtar~
        and~Christine~A.~Shoemaker,~\IEEEmembership{Member,~IEEE}
\thanks{T. Akhtar is with the National University of Singapore Environmental Research Institute (NERI), Singapore, email: erita@nus.edu.sg}
\thanks{C. Shoemaker is with the Department of Industrial Systems Engineering and Management and the Department of Civil and Environmental Engineering, National University of Singapore, Singapore, email: shoemaker@nus.edu.sg}
\thanks{This work was supported in part by the NSF Grant CISE 1116298 to Prof. Shoemaker and NRF Singapore's CREATE program (E2S2-CREATE).}
\thanks{Submitted to IEEE Transactions on Evolutionary Computation, 2019.
}}

\maketitle

\begin{abstract}
Multi-Objective Optimization (MOO) is very difficult for expensive functions because most current MOO methods rely on a large number of function evaluations to get an accurate solution. We address this problem with surrogate approximation and parallel computation. We develop an MOO algorithm MOPLS-N for expensive functions that combines iteratively updated surrogate approximations of the objective functions with a structure for efficiently selecting a population of $N$ points so that the expensive objectives for all points are simultaneously evaluated on $N$ processors in each iteration. MOPLS incorporates Radial Basis Function (RBF) approximation, Tabu Search and local candidate search around multiple points to strike a balance between exploration, exploitation and diversification during each algorithm iteration. Eleven test problems (with 8 to 24 decision variables and two real-world watershed problems are used to compare performance of MOPLS to ParEGO, GOMORS, Borg, MOEA/D, and NSGA-III on a limited budget of evaluations with between 1 (serial) and 64 processors. MOPLS in serial is better than all non-RBF serial methods tested. Parallel speedup of MOPLS is higher than all other parallel algorithms with 16 and 64 processors. With both algorithms on 64 processors MOPLS is at least 2 times faster than NSGA-III on the watershed problems. 
\end{abstract}

\begin{IEEEkeywords}
Expensive objectives, Meta-modeling, Parallel algorithms, Tabu list, Radial Basis Functions.
\end{IEEEkeywords}

%
\IEEEpeerreviewmaketitle

\section{Introduction}
%
%
%
%
\subsection{Background and Motivation}
The last decade has seen a significant increase both in parallel computing and in the application of Multi-Objective Optimization (MOO) by researchers to real world problems, including simulation-optimization problems. MOO has been applied to numerous application areas including engineering disciplines; machine learning and computer science; and water resource management \cite{CoelloCoello2007, Razavi2012, Gong2016, Koziel2014, Perez2017}. 

Many real world  Multi-Objective simulation-optimization problems can be computationally expensive. A single evaluation of objective functions via simulation may require minutes to days \cite{Espinet2013, Horn2016, Kourakos2013} and hence, we want algorithms that can obtain good solutions within a small number of expensive objective function evaluations and/or short wall-clock time.  This study describes a new algorithm for Multi-Objective optimization that is efficient because it uses surrogate approximation as well as a parallel population-based candidate search mechanism.

\subsection{Literature Review}
Research on Multi-Objective Optimization (MOO) algorithms for simulation-optimization has been dominated by Multi-Objective Evolutionary Algorithms (MOEA) \cite{CoelloCoello2007, IMSExample:Zhang2007, IMSExample:Deba}. The population based structure of evolutionary algorithms assists in maintaining convergence and diversity for MOO \cite{debbook}. Additionally, evolutionary algorithms can be easily parallelized using a simple synchronous master-slave framework \cite{Goldberg} \cite{MOEA_water} (described in Section \ref{sec_parallel_paradigm}), and can increase efficiency in solving computationally expensive simulation-optimization problems. Recent contributions in MOEAs have also focused on development of hybrid and multi-method algorithms \cite{Vrugt2007, Hadka2013} and on many-variable and many-objective optimization \cite{Deb2014A, Deb2014B}. Even with considerable developments made in MOEA research, evolutionary algorithms typically require many objective function evaluations and hence, may not be suitable for MOO problems with computationally expensive objective functions. 

Surrogate models (also referred to as response surface models or meta-models) can be effectively incorporated into MOO algorithms, iteratively, to  significantly reduce the number of function evaluations. Such algorithms have been developed for both single-objective optimization (SOO) and MOO. They have been referred to as surrogate algorithms, iterative response surface algorithms or Model-Based Multi-Objective (MBMO) algorithms in prior literature \cite{Horn2015, Akhtar2015, Namura2017, Muller2017}.     

Most surrogate algorithms developed for MOO employ the kriging surrogate \cite{dace} (alternatively called the Gaussian Process Model) and are inspired by the Efficient Global Optimization (EGO) algorithm framework proposed by Jones et. al \cite{IMSExample:Jones1998} for single objective optimization. Some of these algorithms (for instance, ParEGO and MOEA/D-EGO) \cite{IMSExample:Knowles2006, Zhang2010b} aggregate the multi-objective problem into multiple single objective problems . Some algorithms maximize the Expected Improvement (EI) of hypervolume or some other infill criterion during the kriging-assisted search  \cite{IMSExample:Ponweiser2008, Emmerich2016, Namura2017}. Some studies have also coupled EGO with MOEAs \cite{IMSExample:Beume2007, Zhang2010b, IMSExample:Obayashi2005}. Horn et. al. \cite{Horn2015} provide a comprehensive classification of kriging-based MOO algorithms introduced in prior literature. 

Surrogate models other than kriging (or Gaussian Process (GP) models) have also been used in iterative efficient MOO algorithms. The choice of surrogates that have been used in prior literature include Radial Basis Functions (RBFs) \cite{Akhtar2015,  Muller2017, Regis2016, Datta2016, Karakasis2006, Georgo, IMSExample:Santana2007, Martinez2013}, Support Vector Machines (SVMs) \cite{Martinez, Perez2013} and Neural Networks \cite{Deb2007, Kourakos2013}. Most of these algorithms combine surrogate models with MOEAs to propose point(s) for expensive evaluations in each algorithm iteration. Prior studies that use RBFs as surrogate models for both single and multi-objective surrogate optimization have shown that deterministic RBFs could be more effective than GP models (and other surrogates) \cite{Akhtar2015,  Muller2017, Regis2007, Wild2007, Regis2013a, Mueller2013a}. For instance, Akhtar and Shoemaker \cite{Akhtar2015} propose an RBF-based MOO algorithm, GOMORS, that with application to numerous test functions and a groundwater remediation problem, is more efficient than a kriging-based MOO algorithm, ParEGO \cite{IMSExample:Knowles2006}, especially for problems with more than ten decision variables.  

Efficiency of iterative surrogate methods can be further improved if algorithms can propose multiple points for parallel evaluation in each algorithm iteration. Strategies have been proposed in the past for selecting multiple points in an algorithm iteration \cite{Akhtar2015, Horn2015, Zhang2010b, Namura2017} to facilitate parallel computation with each evaluation on a separate processor. Prior research also shows that local search (i.e, heuristic neighbourhood search) can be effective in improving efficiency of both surrogate and non-surrogate MOO algorithms \cite{Akhtar2015, Martinez2013, Martinez, IMSExample:Santana2007, Zhang2010b, Georgo}. Coello et al. (2007) \cite{CoelloCoello2007} discuss the inherent advantages of incorporating local search schemes (for instance, hill climbing, simulated annealing, Tabu search, etc.) in hybrid MOEAs and their success in terms of improving convergence, diversity, robustness and efficiency in Multi-Objective Optimization.

\subsection{Contribution of this Study}
This study introduces a new parallel population-based MOO framework, that incorporates surrogate-assisted local "candidate" search, Radial Basis Functions, adaptive learning and a novel Tabu mechanism for selection of multiple points for expensive evaluation in each algorithm iteration. The objective functions can be multimodal (multiple local minima). Our new multi-objective algorithm is called MOPLS and it is designed  for optimization within a limited budget of evaluations. 

The novel contribution of MOPLS is the method used to generate $N$ new points to evaluate simultaneously with $N$ processors. Within this novel method MOPLS introduces a memory archive that adaptively selects a population of new points for surrogate-assisted local candidate search. Furthermore, MOPLS is designed within a generational population-based synchronous parallel framework, to select (via candidate search) and evaluate a batch of $N$ new points for expensive evaluation in each algorithm iteration. The population-based structure allows MOPLS to work well with a large number of parallel processors and results are given up to 64 processors.

\section{Problem Description}

Let $\mathcal{D}$ (decision space) be a unit hypercube and a subset of $\mathbb{R}^{d}$ and $x$ be a variable in the decision space, i.e., $x \in [0,1]^d$ (Any hyperrectangular domain can be mapped into the unit hypercube). Let the number of objectives equal $k$ and let  $F(x) = \{ f_{1}(x),\dots,f_{k}(x) \}$ be the vector of objectives such that $F(x) \in \mathcal{F}$ (objective space). So $f_j$ denotes the j\textsuperscript{th} objective, where $f_j$ is a function of $x$ and $f_j \!  : \!  \mathcal{D}  \! \mapsto  \! \mathbb{R}$ for $1 \! \le  \! j \! \le \! k$. The framework of the Multi-Objective Optimization problem we aim to solve is as follows:
 
 \begin{equation}
 \begin{split}
 \mbox{minimize   }&   F(x) = [ f_1(x),\dots,f_k(x)]^T \\
 \mbox{subject to   }&   x \in [0,1]^d
 \end{split}   
 \end{equation} 

  
  The focus of this study is on employing a \textit{posteriori} \cite{Cohon1975} preference optimization, where the purpose of the multi-objective problem is to find a set of Pareto-optimal (or trade-off) solutions,  $ P^*=\{ x^{*} \mid x^{*} \! \in \! \mathcal{D} \}$.
 \newtheorem{mydef}{Definition} 
 \begin{mydef}
.  A solution $x_{1} \! \in \! \mathcal{D}$  \textbf{dominates} another solution $x_{2} \! \in \! \mathcal{D}$ ($x_1 \! \preceq \! x_2$) if and only if $f_{j} (x_{1}) \le f_{j} (x_{2})$ for all $1\! \le \! j \! \le \! k$, and  $f_{j} (x_{1}) < f_{j} (x_{2})$ for some $1\! \le \! j \! \le \! k$.
\end{mydef}

\begin{mydef}
 Given a set of solutions $S= \{ x \mid x \! \in \! \mathcal{D} \}$ , a subset of solutions $S^\dagger \subset S$ is \textbf{non-dominated} in $S$ if there does not exist a solution $x \! \in \! S$ which dominates $x^\dagger \! \in \! S^\dagger$, i.e, $S^\dagger = \{ x^\dagger \! \in \! S \mid  \not\exists  x \! \in \! S, x \!  \preceq \!  x^\dagger \}$.
 \end{mydef}
 
 \begin{mydef}
 A set of solutions $ P^* \! \subset \! \mathcal{D} $ which is non-dominated in the decision space, i.e, $\mathcal{D}$ is called a \textbf{Pareto optimal set}. Hence, $P^* = \{ x^* \! \in \! \mathcal{D} \mid  \not\exists  x \! \in \! \mathcal{D}, x \!  \preceq \!  x^* \}$.  
 \end{mydef}

As per the above definitions, the aim of a MOO algorithm is to find the set of Pareto optimal solutions $ P^*$. The set of objective vectors corresponding to $P^*$ is called the the \textit{Pareto front}. Since computation of the objective function vector is expensive, the aim of our analysis is to identify a set which closely approximates $ P^*$, within a limited number of function evaluations.



\section{Surrogate-assisted - Multi-Objective Parallel Local Search (MOPLS)}


Multi Objective Population-based Parallel Local Surrogate-assisted search, i.e. MOPLS, is an iterative surrogate algorithm designed for computationally expensive Multi-Objective (MO) blackbox optimization problems. Iterative surrogate algorithms are also referred to as Sequential model-based optimization (SMBO) methods in prior literature \cite{Hutter2011}.  

An illustration of the general algorithm framework of MOPLS is provided in Figure \ref{fig_framework}, where Step 2 is the iterative step. MOPLS employs Radial Basis Functions (RBF) as iterative surrogates to improve efficiency and speed of convergence. Algorithm  efficiency is improved further via paralleization of the iterative step (see Step 2 in Figure \ref{fig_framework}), where simultaneous surrogate-assisted local \textit{candidate searches} (discussed further in Section \ref{ch4_local}) are performed (in Step 2.2, see Figure \ref{fig_framework}) around selected (in Step 2.1, see Figure \ref{fig_framework})  previously evaluated points, called \textit{center points} or \textit{center point population}. The parallel iterative framework of MOPLS is similar to the generational framework \cite{CoelloCoello2007} of Multi Objective Evolutionary Algorithms (MOEAs), since a parent population of "centers" is selected in each MOPLS iteration, and a new population (or batch of points) is subsequently selected and evaluated via candidate search on the parent population of centers. 

\subsection{Synchronous Master-Salve Parallelization}
\label{sec_parallel_paradigm}

\begin{figure}
\centering
 \noindent\includegraphics[ width = 21pc]{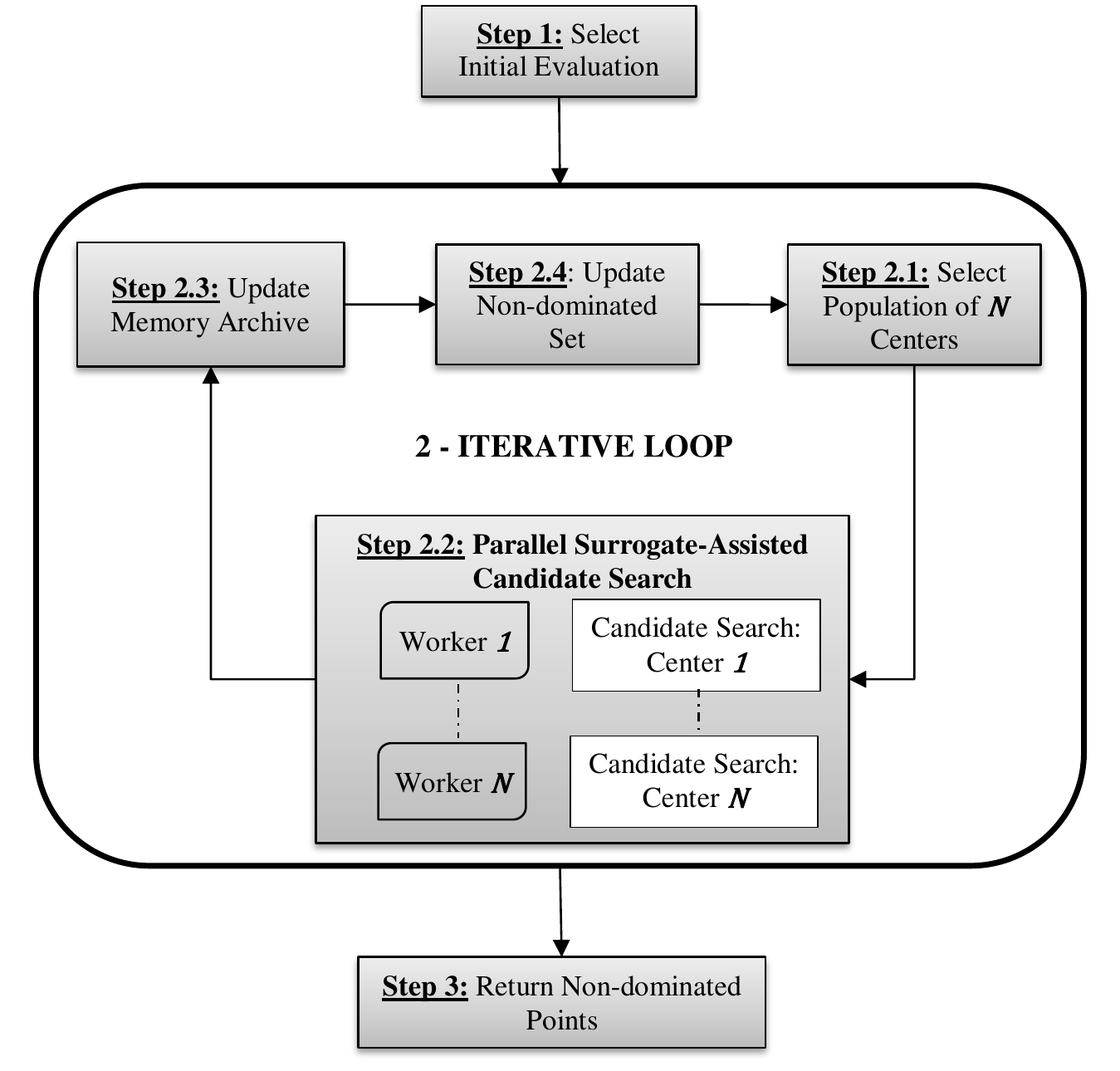}
 \caption{Schematic view of the MOPLS Alogrithm Framework.}
 \label{fig_framework}
 \end{figure}

A synchronous Master-slave architecture (also popular for generational MOEAs \cite{Hadka2015}) is employed for parallelization of MOPLS. Within the iterative loop of MOPLS (i.e., Step 2 in Figure \ref{fig_framework}), the Master process first selects a population of $N$ centers (Step 2.1 in Figure \ref{fig_framework}), from the set of points that have already been expensively evaluated, and sends each center point to a slave process (or 'worker' as referred to in Figure \ref{fig_framework}).The number of centers, i.e., $N$ (also referred as population size), is an algorithm parameter (see Table \ref{table:params} for list of MOPLS paramaters). Moreover, we assume that there are $N$ (as depicted in Step 2.2 of Figure \ref{fig_framework}) workers available. Hence, each worker performs an independent surrogate candidate search (described in detail in Sections \ref{ch4_frame} and \ref{ch4_local}) around the center point assigned to it, to select a new point for evaluation, and also evaluates this new point. (This is Step 2.2 in Figure \ref{fig_framework}.) The master process waits for all $N$  workers to finish their selection and evaluation tasks and then executes Step 2.3 and 2.4 (described in detail in Sections \ref{ch4_frame} - \ref{ch4_local}) of the iterative loop.  

The synchronous Mater-slave parallel framework of MOPLS does not require derivative information and is suitable for expensive functions that can be multi-modal. 

\begin{table*}[ht!]
\caption{Definitions of Sets and Variables of MOPLS} 
\centering 
\resizebox{\textwidth}{!}{
\begin{tabular*}{\textwidth}{ l ll } 
\hline\hline  
\bf{Item} & \bf{Description} \\ [0.5ex] 
\hline \\ 
$F(x)= [f_{1},\dots,f_{k}]$ & Expensive objectives for the multiple-objective optimization problem.  \\ [0.2ex] 
$z_i = (x_i, y_i, r_i, c_i, c^{'}_{i})$ & A multi-attribute element where $x_i \in \mathbb{R}^{d}$ is the decision variable, $y_i \in \mathbb{R}^{k}$ is the objective vector corresponding to $x_i$, i.e.,  \\ &   $y_i = F(x_i)$, $r_i \in \mathbb{R}$ is the local search radius of $x_i$, and $c_i$ and $c^{'}_{i}$ are the failure count and tabu count, respectively, of $x_i$. \\ [0.2ex] 
$S_m = \{ z_1, \dots, z_m \}$  & Multi-attribute ordered set corresponding to all points, $\{ x_1, \dots, x_m \}$,  which have been evaluated via costly simulation \\ & before and including iteration $m$ of the algorithm. \\ [0.2ex] 
$I_{cen} = \{ ind_1 \dots ind_{N} \}$ & The index set corresponding to the evaluated set $S_m$, that indexes the subset of evaluated points that are selected \\ & as the population of center points in Step 2.1 of MOPLS ($|I_{cen}| = N$).  \\ [0.2ex]
$S_{cen} = \{ z_1, \dots, z_{N} \}$ & The subset of evaluated points ($S_{cen} \subset S_m$) chosen as center points in Step 2.1, i.e., $S_{cen} = \! = \! \{z_i \! \in \! S_m \! \mid \!  i \! \in \! I_{cen}\}$.  \\ [0.2ex] 
$z^{*}_i = (x^{*}_i, y^{*}_i, r^{*}_i, c^{*}_i, c^{*'}_{i})$ & A multi-attribute element, where $x^{*}_i$ is the point selected for expensive evaluation after surrogate-assisted local search \\ & around the center point, $x_i$, in Step 2.2 ($x_i$ is an attribute of $z_i \in S_{cen}$), $y_i^{*} = F(x_i^{*})$, $r^{*}_i = r_{init}, c^{*}_i = 0$, and $c^{*'}_{i} = 0$. \\ [0.2ex] 
$S_{new} = \{ z^{*}_1 \dots z^{*}_{N}\}$ & Multi-attribute ordered set of all evaluated points obtained from the parallel surrogate-assisted local search in Step 2.2.  \\ [0.2ex] 
$S_{tabu}$ &  The subset of evaluated points ($S_{tabu} \subset S_m$) which are in the Tabu archive. The Tabu archive is updated in Step 2.3. \\ [0.2ex] 
$P_{m}=\{ z \! \in \! S_m \}$ & The \textit{non-dominated} solutions in $S_m$ based on expensive evaluations.  \\ [0.2ex]
$H_V(Y)$ & Hypervolume of the set of objective function vectors $Y = \{F(x_1), \dots, F(x_n)\}$ is the volume of the objective space \\ & dominated by $Y$. The objective space is bounded by a reference vector, $b$ (see Figure \ref{ch4_hyp_visual}(a) for illustration).  \\ \\
\hline 
\end{tabular*}}
\label{table:nonlin} 
\end{table*}

\subsection{General Algorithm Description} 
\label{ch4_frame}

The MOPLS algorithm is divided into three core steps, namely i) initialization (i.e., Step 1 in Figure \ref{fig_framework}), ii) iterative improvement (i.e., Step 2 in Figure \ref{fig_framework}) and iii) termination (i.e., Step 3 in Figure \ref{fig_framework}). While Figure \ref{fig_framework} provides a schematic overview, a more detailed description of the core steps of MOPLS is provided below. The sets and variables used in the following description of MOPLS are defined in Table \ref{table:nonlin}. Moreover, the search mechanics of MOPLS is dependent upon numerous algorithm parameters (that are also mentioned in the algorithm description below) which are listed and defined in Table \ref{table:params}. Description of MOPLS is as follows:


\noindent\hrulefill \\
\textbf{Step 0 - Define Algorithm Parameters:} \\
$E_T$ - Maximum number of expensive function evaluations\\
$E_I$ - Number of initial expensive evaluations\\
$r_{init}$ - The initial local search radius (discussed in Step 2.2 of the algorithm) \\
$N$ -  Number of \textit{center points} for simultaneous surrogated-assisted search which is usually also the number of processors. \\\\
\textbf{Step 1 - Initial Evaluation Points Selection:} The master process is initialized, and selects, using an experimental design, an initial set (ordered) of points $\{x_{1}, \dots ,x_{m}\}$, where $x_{i} \! \in \! \mathcal{D}$, for $1 \! \le \! i \! \le \! m$, and $m = E_I$. The master initiates $N$ workers and evenly distributes function evaluation tasks to the workers. The workers evaluate the objective  $F(x)= [f_{1}, \dots ,f_{k}]$ at the selected $E_I$ points, via expensive simulations, and return results to the master. Let $\{ y_1, \dots ,y_m \}$ be the objective evaluations (returned by the workers to the master process) corresponding to $\{x_{1}, \dots ,x_{m}\}$, i.e, $y_i = F(x_i)$ for $1 \! \le \! i \! \le \! m$. 

After initial evaluations are done, the master initiates the multi-attribute ordered set, $S_m = \{ z_1, \dots, z_m \}$, of all points that have been expensively evaluated so far (see Table \ref{table:nonlin} for definition of sets). Each element, $z_i$ (see Table \ref{table:nonlin} for definition), of the set $S_m$ has five attributes, i.e., $z_i = (x_i, y_i, r_i, c_i, c^{'}_{i})$, where $x_i$ is the i\textsuperscript{th} expensively evaluated point, $y_i$ is corresponding the expensive objective function vector, $r_i = r_{init}$, $c_i = 0$, and $c^{'}_{i} = 0$ (Note: $r_i$, $c_i$ and $c^{'}_{i}$ are called memory attributes that are updated in Step 2.3). The master also initializes $S_{tabu} = \{ \}$ as an empty set, where $S_{tabu}$ is the set of evaluated points that cannot be selected in the population of \textit{center points} in the iterative framework of the algorithm (Step 2 below). Let $P_{m} \! = \!\{ z_i \! \in \! S_m  \mid x_i \mbox{ is non-dominated in } \{x_{1}, \dots ,x_{m}\}\}$  be the set of non-dominated points from $S_{m}$.   \\\\ 
\textbf{Step 2 - Iterative Improvement:} Run algorithm iteratively until termination condition is satisfied:\\
\textbf{While $m \le E_T$}
\setlength\parskip{0in}
\begin{itemize}
 \item[] \textbf{Step 2.1 -Select Center Points:}  The master selects $N$ center points from $S_{m}$ based on non-dominated sorting and hypervolume contributions (details discussed in Section \ref{ch4_center} and \textit{Boxed Algorithm 1}). Let $I_{cen} = \{ ind_1 \dots ind_{N} \}$ be the indices of the selected centers corresponding to the ordered set $S_m$.
\item[] \textbf{Step 2.2 - Parallel Surrogate-Assisted Local Candidate Search:} Let $S_{cen}$ (see Table \ref{table:nonlin} for definition) denote the set of $N$ evaluated points chosen as center points in Step 2.1, i.e, $S_{cen} \! = \! \{z_i \! \in \! S_m \! \mid \!  i \! \in \! I_{cen}\}$. The master initiates $N$ worker processes and sends a center point, $z_i \in S_{cen}$ to each worker, $i$, where $1 \! \le i \! \le N$:\\
\textbf{For $i = 1, \dots, N$} (Parallel Execution)
\setlength\parskip{0in}
\begin{itemize}
\item[] Each worker process, $i$, performs an independent parallel Surrogate-Assisted Local Candidate Search (details given in Section \ref{ch4_local} and \textit{Boxed Algorithm 2}) around the center point $ z_i \in S_{cen}$ and selects (and evaluates) a new point, $x_i^*$, for expensive evaluation. The worker, $i$, returns $z^{*}_i$ to the master process. Note that $z^{*}_i = (x^{*}_i, y^{*}_i, r^{*}_i, c^{*}_i, c^{*'}_{i})$ is a multi-attribute element, where $x^{*}_i$ is the point selected for expensive evaluation, $y_i^{*} = F(x_i^{*})$, and $r^{*}_i, c^{*}_i$, and $c^{*'}_{i}$ are memory attributes corresponding to $x^{*}_i$.
\end{itemize}
\textbf{End Loop} \\
$S_{new} = \{ z^{*}_1 \dots z^{*}_{N}\}$ is the ordered set of all new evaluated points obtained (by the master) from the parallel surrogate-assisted local search of the current iteration. 
\item[] \textbf{Step 2.3 - Update Tabu List and Memory Attributes:} The master updates the tabu list, i.e, $S_{tabu}$. The master also updates memory attributes, i.e, $r_i$, $c_i$ and $c^{'}_i$, of each evaluated point $z_i = (x_i, y_i, r_i, c_i, c^{'}_{i})$, where $z_i \in S_m$. The tabu list and memory attribute update procedure is discussed in Section \ref{ch4_tabu} and \textit{Boxed Algorithm 3}. 
\item[] \textbf{Step 2.4 - Update the evaluated set and the non-dominated set:} The set of evaluated points is updated , i.e, $S_{m} = \{S_{m} \} \cup  \{ S_{new} \}$ and $m = m + N$. Update the non-dominated set $P_m$, i.e, compute $P_{m} \! = \!\{ z_i \! \in \! S_m  \mid x_i \mbox{ is non-dominated in } \{x_{1}, \dots ,x_{m}\}\}$.
\end{itemize}
\textbf{End Loop} \\\\
\textbf{Step 3 - Return Best Approximated Front:}   Return $P_{m} \! = \!\{ z_i \! \in \! S_m  \mid x_i \mbox{ is non-dominated in } \{x_{1}, \dots ,x_{m}\}\}$ as an approximation  to the Pareto optimal set.

\noindent \hrulefill \\
\setlength\parskip{0in}

The algorithm initialization phase (Step 1) starts off with sampling of $E_I$ points $x_1 \dots x_{E_I}$ from decision space, $\mathcal{D}$, via an experimental design method. We use Latin hypercube sampling to generate the initial sample points \cite{Mckay1979}.  Expensive simulation is employed to evaluate the values of all objectives for the initial sample points. 

Step 2 constitutes the iterative framework of MOPLS and is the core component of the algorithm. Each MOPLS iteration begins with selection of \textit{center points} (Step 2.1), which is based on ranking of evaluated points in terms of their convergence and diversity potential. The selection of \textit{center points} is analogous to selection of the parent population in a generational \cite{CoelloCoello2007} evolutionary algorithm. Methodology for selection of center points is discussed in detail in Section \ref{ch4_center} and \textit{Boxed Algorithm 1}.

Selection of center points (or parent population) is followed by simultaneous surrogate-assisted local searches (Step 2.2) around each center point within the parent population. As discussed in the above algorithm description and in Section \ref{sec_parallel_paradigm}, Step 2.2 (also see Figure \ref{fig_framework}) of MOPLS is executed in Parallel. Details of the local candidate search methodology of Step 2.2 are provided in Section \ref{ch4_local} and \textit{Boxed Algorithm 2}.  

A memory archive and tabu list are also maintained, and updated (Step 2.3) in each iteration of MOPLS. The memory archive adaptively changes the local search neighborhood of the evaluated points. The tabu list is a secondary memory archive, which prohibits the inclusion of certain evaluated points in the population of center points. The procedure of maintaining and updating the memory archive and tabu list is discussed in detail in Section \ref{ch4_tabu} and \textit{Boxed Algorithm 3}.

\begin{table}
\caption{Definitions of Algorithm Parameters} 
\centering 
\begin{tabular}{ c l } 
\hline\hline  
\bf{Item} & \bf{Description} \\ [0.3ex] 
\hline \\ 
$E_I$ & The number of initial expensive evaluations.  \\ [0.2ex] 
$E_T$ & The number of total expensive evaluations. \\ [0.2ex] 
$N$  & The number of \textit{center points} (assumed to be equal to the no. of \\ &  available processors)for simultaneous local surrogate-assisted \\ &  candidate search in each algorithm generation / iteration. \\ [0.2ex] 
$r_{init}$ & Parameter that controls the radius of local search neighborhood. \\ \\ 
\hline 
\end{tabular}
\label{table:params} 
\end{table}


\subsection{Selection of Center Points}
 \label{ch4_center}

Selection of center points, i.e., Step 2.1 (see Figure \ref{fig_framework}) is the first step of each iteration (Step 2; See Figure \ref{fig_framework}) of MOPLS. The set of center points (also called parent population), denoted as denoted as $S_{cen}$ is selected from the set of all points evaluated so far, i.e., $S_m$ (see Table \ref{table:nonlin} for definitions). Hence, $S_{cen} \subset S_m$. Furthermore, the number of points in $S_{cen}$ is $N$, where $N$ is an algorithm parameter (see Table \ref{table:params}) that denotes the number of center points (or population size) chosen for parallel local search in each algorithm iteration.   

\begin{figure}
\centering
 \noindent\includegraphics[ width = 21pc]{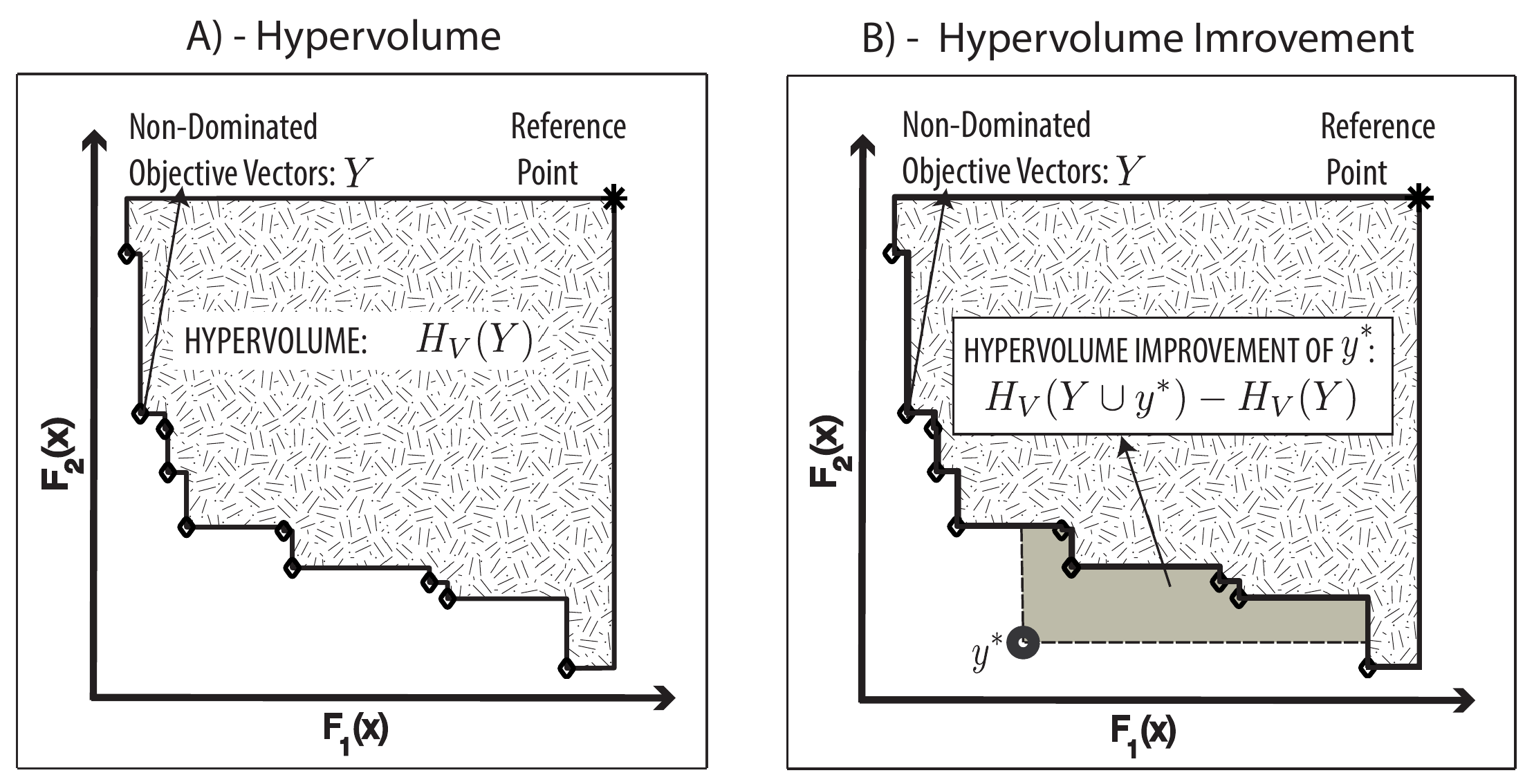}
 \caption{Visualization of a) Hypervolume (hatch pattern) and b) Hypervolume Improvement (grey fill color) employed in Step 2.2 for selection of new evaluation points (discussed in Section \ref{ch4_local}).}
 \label{ch4_hyp_visual}
 \end{figure}

The detailed algorithmic framework of the center selection step, i.e, Step 2.1 (see Figure \ref{fig_framework}), of MOPLS is provided in \textit{Boxed Algorithm 1}. As depicted in \textit{Boxed Algorithm 1}, selection of centers starts with initiation of the set of center points, i.e., $S_{cen}$, as an empty set (see line 4), and points are iteratively added to $S_{cen}$ (from set of evaluated points, i.e., $S_m$) until there are $N$ points in $S_{cen}$ (lines 5-20).

\begin{figure}
 \centering
 \noindent\includegraphics[ width = 21pc]{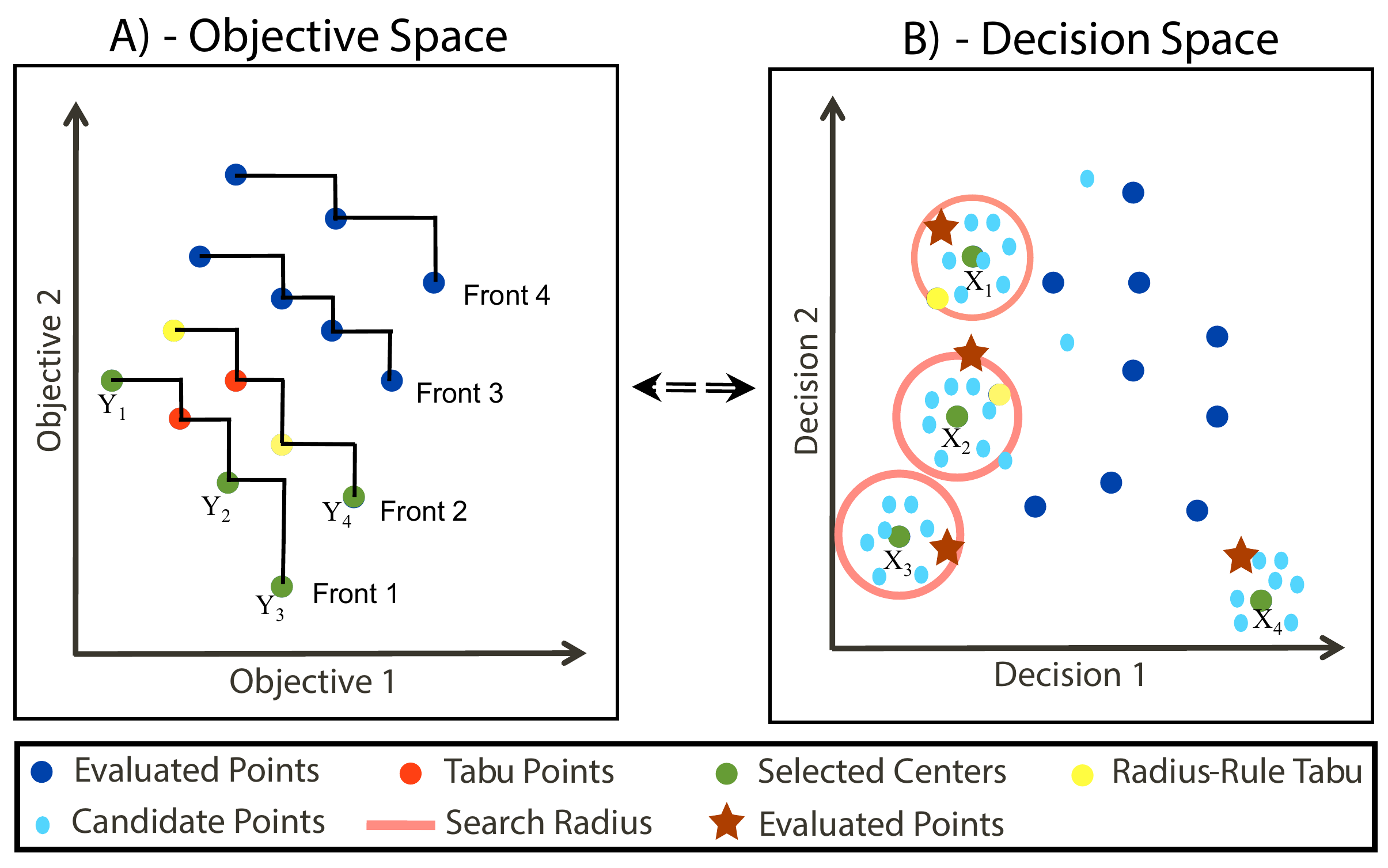}
 \caption{Illustration of a) How centers are chosen in Step 2.1 of MOPLS, and b) How new points are selected around centers via candidate search of Step 2.3 of MOPLS (Assumed here that number of centers is 4).}
 \label{center_visual}
 \end{figure}

The iterative selection of center points (as depicted in \textit{Boxed Algorithm 1}) follows a two-layered strategy to rank evaluated solutions for selection. The first layer of ranking is based on the  \textit{non-dominated sorting} algorithm \cite{Goldberg} \cite{debbook}. The {\it{non-dominated sorting}} algorithm computes the non-dominated solutions in the evaluation set ($S_{m}$), assigns them the highest rank (or 'Front number'), removes them from contention, and assigns the next rank to the non-dominated solutions amongst the remaining evaluated solutions. This process is typically repeated to rank all solutions into a number of fronts. (See Figure \ref{center_visual}(a) for visual illustration of non-dominated sorting.) However, in MOPLS (see lines 7 and 20 of \textit{Boxed Algorithm 1}), we continue to find non-domination ranks (or non-dominated fronts) only until $N$ solutions are added to the set of center points (line 5).  

The selection of center points in MOPLS employs a second layer of ranking in order to differentiate evaluated points that have the same non-domination ranking. In \textit{Boxed Algorithm 1} (see Line 7) the set $S^*$ is the subset of evalauted points that are on a non-dominated front. We use the hypervolume (also called $S$-Metric) contribution metric \cite{IMSExample:Ponweiser2008} \cite{Zitzler2007} to rank and sort each point within the set $S^*$ (see lines 8-9 of \textit{Boxed Algorithm 1}), before points from $S^*$ are selected and added to the set of center points $S_{cen}$. The hypervolume contribution of a point $y$ in the set $S^*$ is the difference in hypervolume (see Table \ref{table:nonlin} and Figure \ref{ch4_hyp_visual}(a) of definition of hypervolume) of $S^*$ and hypervolume of $S^*$ excluding the point $y$. Hence, higher values of the hypervolume contribution are desirable. (See the procedure referenced in Line 8 of \textit{Boxed Algorithm 1} for definition of hypervolume contribution.)




\begin{algorithm}
\fontsize{9.2}{7.2}\selectfont
\LinesNumbered
\NoCaptionOfAlgo
\SetKwData{Left}{left}
\SetKwData{This}{this}
\SetKwData{Up}{up}
\SetKwFunction{HV}{HypervolumeContributions}
\SetKwFunction{FindNonDominated}{\text{FindNonDominated}}
\SetKwFunction{Sort}{SortFront}
\SetKwInOut{Input}{Input}
\SetKwInOut{Output}{Output}
\BlankLine
\Input{1) $N$ - Number of centers per generation.\\  2) $m$ - Number of points evaluated so far.\\ 3) $S_{m}$ - Set of evaluated points. \\4) $S_{tabu}$ - Set of \textit{tabu} evaluated points.
\\
5) $E_{I}$ - Number of initial expensive evaluations. \\
6) $E_{T}$ - Number of total expensive evaluations. }
\BlankLine
\Output{1) $I_{cen}$ - Indices of the selected centers.}
\Begin{
$S = S_m$\;
$d_{thresh} = 1 - \frac{m - E_I}{E_T - E_I}$\;
$S_{cen} = \{\}$\;
\While{$|S_{cen}| \le N$}{
$X = \{x_i \mid (x_i, y_i, r_i, c_i, c^{'}_i) \! \in \! S, 1 \! \le \! i \! \le \! |S| \}$ \;
$S^* = \{ (x_i, y_i, r_i, c_i, c^{'}_i) \! \in \! S  \mid x_i \mbox{ is non-dominated in } X \}$\; 
$HC$ =  \HV{$S^*$}\; 
$ S^{*}$ = \Sort{$S^*,HC$}\; 
\For{$i = 1$ to $|S^*|$ {\bf and} $(x_i, y_i, r_i, c_i, c^{'}_i) \in S^*$}{
Let $z_i = (x_i, y_i, r_i, c_i, c^{'}_i)$ \;
\If{$z_i \not\in S_{tabu}$}{
\If{$|S_{cen}| == 0$}{
$S_{cen} = \{S_{cen}\} \cup \{z_i\}$\;}
\ElseIf{$|S_{cen}| \le N$}{
\For{$j = 1$ to $|S_{cen}|$ {\bf and} $(\hat{x}_j, \hat{y}_j, \hat{r}_j, \hat{c}_j, \hat{c}^{'}_j) \in S_{cen}$}{
\If{$||\hat{x}_j - x_i|| \le \hat{r}_j*d_{thresh}$}{
go to line 9\;}}
$S_{cen} = \{S_{cen}\} \cup \{z_i\}$\;}
} 
}
$S = \{S\} \backslash \{S^*\}$\;}
$I_{cen} = \{i \mid a_i \! \in \! S_m, b_j \! \in \! S_{cen}, a_i = b_j, 1 \! \le \! j \! \le \! N \}$\;
}
\BlankLine
 \setcounter{AlgoLine}{0}
\SetKwProg{myproc}{Procedure}{}{}
 \myproc{\HV{$S^*$}}{
 $Y^* = \{y_i \mid (x_i, y_i, r_i, c_i, c^{'}_i) \! \in \! S^*, 1 \! \le \! i \! \le \! |S^*| \}$ \;
\For{$i = 1$ to $|Y^*|$ {\bf and} $y_i \in Y^*$}{
$HC_i = \bigg[ H_V(Y^*) - H_V(\{Y^*\} \backslash \{y_i\}) \bigg]$\;
}  
 \KwRet\ $HC$;}
 \BlankLine
  \setcounter{AlgoLine}{0}
 \myproc{\Sort{$S^*,HC$}}{
Sort (descending order) the elements of $S^*$ according to the hypervolume contribution values recorded in $HC$\;
 \KwRet $S^*$\;}
\caption{\textbf{Boxed Algorithm 1: Center Selection}}
\end{algorithm}
 
MOPLS also maintains a list of (dynamically changing) tabu solutions ($S_{tabu}$), which cannot be selected as centers. The procedure for updating the tabu list is discussed in Section \ref{ch4_tabu} and \textit{Boxed Algorithm 3}. Figure \ref{center_visual}(a) provides an illustration of how the population of centers is selected in MOPLS, assuming that number of centers is 4. Figure \ref{center_visual}(a) depicts the tabu solutions in red, and the center selection algorithmic framework incorporates the tabu list within the selection mechanism (see lines 10-12 of \textit{Boxed Algorithm 1}). 
 
 In addition to consideration of tabu status of previously evaluated points, the iterative center selection strategy of MOPLS also ensures that a point that is close to another selected center (that is already selected in the current iteration) is not considered for selection (see lines 15-18 of \textit{Boxed Algorithm 1}). This selection rule is called the "radius-rule". Figure \ref{center_visual} illustrates that there are two points that are close to already selected centers for the current iterations (depicted in yellow), and hence, are not selected as centers. The closeness is deduced by a threshold, $d_{thresh}$ (where $d_{thresh} = 1 - \frac{m - E_I}{E_T - E_I}$), multiplied by the local search radius $r_{i}$ (see line 17 of \textit{Boxed Algorithm 1}). At the beginning of the iterative loop of MOPL the value of $d_{thresh}$ is equal to 1. Hence, points that lie within the search neighborhoods of already selected center points, are not considered for selection as centers. 
 
 However, the value of $d_{thresh}$ is reduced to 0, as the algorithm progresses, in order to allow points that are somewhat close to each other in the decision space to be selected as center points simultaneously. This enables more concentrated local search in regions of interest in the decision space as the number of evaluations increase.

\subsection{Parallel Surrogate-Assisted Local Candidate Search}
\label{ch4_local}


 Step 2.2 (see Figure \ref{fig_framework}) of MOPLS employs a parallel surrogate-assisted local search to choose new points for expensive evaluations. As is depicted in the general algorithm framework of Figure \ref{fig_framework} and explained in Sections \ref{sec_parallel_paradigm} and \ref{ch4_frame}, the master process sends one point each from the set (or parent population) of $N$ center points to each worker in Step 2.2. Each worker, $i \in {1, \dots, N}$ independently (and in parallel) selects a new point for expensive evaluation after performing a surrogate-assisted search in the local neighborhood of the center point sent by the master process (denotes as $z_i$ ,i.e., input no. 2 in \textit{Boxed Algorithm 2}). The worker process subsequently evaluates the selected point via expensive evaluation and sends the evaluated point back to the master process. The procedure adopted by each worker, $i$, in selection and evaluation of the new point is depicted in \textit{Boxed Algorithm 2} and discussed below.  
 
\begin{algorithm}
\fontsize{8}{6}\selectfont
\LinesNumbered
\NoCaptionOfAlgo
\SetAlgoSkip{smallskip}
\SetAlCapSkip{0ex}
\SetKwData{Left}{left}
\SetKwData{This}{this}
\SetKwData{Up}{up}
\SetKwFunction{GC}{GenerateCandidates}
\SetKwFunction{FS}{FitRbfModel}
\SetKwFunction{ES}{EvaluateSurrogate}
\SetKwFunction{FindNonDominated}{FindNonDominated}
\SetKwFunction{HV}{BestHypervolume}
\SetKwFunction{mutate}{mutate}
\SetKwFunction{rand}{Rand}
\SetKwFunction{sims}{Simulate}
\SetKwFunction{mm}{MaxMinMethod}
\SetKwInOut{Input}{Input}
\SetKwInOut{Output}{Output}
\BlankLine
\Input{1) $S_{m}$ - Set of evaluated points. \\2) $z_{i} \in S_{cen}$ - Center point sent to Worker $i$ for surrogate \\ candidate search. Note that $z_{i} = (x_{i}, y_{i}, r_{i}, c_{i}, c^{'}_{i})$  \\ 3) $P_{m}$ - Set of {\it{non-dominated}} evaluated points. $P_{m} \subset S_{m}$. }  
\BlankLine
\Output{$(x^*_i, y^*_i, r^*_i, c^*_i, c^{*'}_i)$ - Point selected for evaluation by Worker $i$ and evaluated via expensive simulation.  }
\Begin{
$p_1 = $ generate a uniformly distributed real number in [0, 1]\;
\If (perform surrogate-assisted search){$p_1 \le prob_{cand}$}{
$\widehat{F} = $ \FS{$x_{cen}$, $S_m$}\;
$X_{cand} = $ \GC{$x_{cen}$, $r_{cen}$}\;
$X^{*}_{cand} = \{ x \! \in \! X_{cand}  \mid x \mbox{ is non-dominated in } X_{cand} \mbox{ as per  } \widehat{F} \}$\; 
$p_2 = $ generate a uniformly distributed real number in [0, 1]\;
\If {$p_2 \le prob_{hv}$}{
$ x^*_i = $ \HV{$X^*_{cand}$, $P_{m}$}\;} 
\Else{
$ x^*_i = $ \mm{$X^*_{cand}$, $S_{m}$}\;} 
}
\Else(mutate center){
$x^*_i = $ \mutate{$x_{cen}$} \;}
$y^*_i = F(x^*_i)$, $r^*_i = r_{init}$, $c^*_i = 0$, $c^{*'}_i = 0$ \;  
}
\BlankLine
\setcounter{AlgoLine}{0}
\SetKwProg{myproc}{Procedure}{}{}
\myproc{\FS{$x_{i}$, $S_m$}}{
Choose a subset, $S$, from already evaluated solutions, $S_m$, which are closest to $x_{i}$, as per euclidean distance \;
Fit response surface models (RSM) for each objective based on a set $S$. Let $\widehat{F}(x)=[ \widehat{f}_{1}, \dots ,\widehat{f}_{k}]$ denote the inexpensive RSMs \;
 \KwRet $\widehat{F}$\;}
\BlankLine
\setcounter{AlgoLine}{0}
\myproc{\GC{$x_{i}$, $r_{i}$}}{
$X_{cand} = \{ \}$, $p = $ generate a uniformly distributed real number in [0, 1]\;
\If ( */ perform hyper-spherical local search */) {$p \le \frac{1}{2}$}{
$\sigma$ = $r_{i}$ \;} 
\Else ( /* perform hyper-eliptical local search */){
$ \sigma = |a|$, where $ a \sim \mathcal{N}(\sigma,\sigma^2/4)$ and $\mathcal{N}$ denotes the normal distribution \; }
$N_{cand} = 500*d$, where $d$ is the number of decision variables\;
\For (generate candidate points) {$j = 1$ to $N_{cand}$}{
$x_{cand,j} = x_{i} + z$ where $z$ is a random vector s.t $z \sim \mathcal{N}(0,\sigma^2I_d)$\;
$X_{cand} = \{X_{cand}\} \cup \{x_{cand,j}\}$ \;
}  
 \KwRet $X_{cand}$ \;}
\BlankLine
\setcounter{AlgoLine}{0}
 \myproc{\HV{$X^{*}_{cand}$, $P_{m}$}}{
 $Y_m = \{y_j \mid (x_j, y_j, r_j, c_j, c^{'}_j) \! \in \! P_m, 1 \! \le \! j \! \le \! |P_m| \}$ \;
$x^*_i = \argmax_{x_j \in X^{*}_{cand}}  \bigg[ H_V(Y_m \cup \widehat{F}(x_j)) - H_V(Y_m) \bigg]$\;
 \KwRet\ $x^*_i$;} 
 \BlankLine
   \setcounter{AlgoLine}{0}
 \myproc{\mm{$X^{*}_{cand}, S_m$}}{
 $X_m = \{x_j \mid (x_j, y_j, r_j, c_j, c^{'}_j) \! \in \! S_m, 1 \! \le \! j \! \le \! |S_m| \}$ \;
$x^{*}_i = \argmax_{x_{j} \in X^{*}_{cand}} \left[ \min_{x_l \in X_m} \| x_{j} - x_{l} \| \right]$\;
 \KwRet $x^*_i$\;}
 \BlankLine
   \setcounter{AlgoLine}{0}
 \myproc{\mutate{$x_{i}$}}{
Generate a point $x^*_i \in \mathcal{D}$ via either \textit{gaussian} or \textit{uniform} mutation of  $x_{i}$ (Mutation probability is $\frac{1}{|x_{i}|}$.)\;
 \KwRet $x^*_i$\;}
\caption{\textbf{Boxed Algorithm 2: Surrogate Search (Worker $i$)}}
\end{algorithm}

\subsubsection{ Response Surface Model}
\label{ch4_rbf}
The first major task of the worker is approximation of the costly functions, $F$, via the inexpensive surrogates, $\widehat{F}(x) = [ \widehat{f}_{1}, \dots ,\widehat{f}_{k} ]$ (line 4 of \textit{Boxed Algorithm 2}). Independent surrogates are fitted for each objective function, $f_j$. 

 Various approximation methods, including artificial neural networks \cite{ANN}, Support Vector Machines (SVM) \cite{svm},  kriging  \cite{dace}, and radial basis functions (RBFs) \cite{Buhmann2003}\cite{Powell90} could be employed to fit the response surface models. Some prior studies \cite{Regis2013a} \cite{Diaz-Manriquez2011} demonstrate the relative effectiveness of RBF approximation in tackling higher dimensional problems (approximately defined as problems with more than 10 decision variables). Muller and Shoemaker \cite{Muller2014} compared surrogate models and found Cubic RBF outperformed kriging and a regression surface (MARS). The worker process of MOPLS  hence employs RBFs as the surrogate modeling methodology for approximating the costly functions.

The training set (denoted as $S$) used for fitting the RBF model is a subset of expensively evaluated points (denoted as $S_m$). The training set, $S$ (see Procedure 'FitRbfModel' referenced at line 4 of \textit{Boxed Algorithm 2}), includes up to 500 points which are closest (as per Euclidean distance) to the center point (denoted as $x_{i}$) in the decision space. The size of the training set is kept limited to avoid excessive computational burden of fitting the RBF model. Hence, the surrogate model is referred as local surrogate model in subsequent discussions. 

\subsubsection{ Generation of Random Candidate Points}
\label{sec_candidates}
Fitting of the local surrogate model is followed by random generation of numerous points around the \textit{center point i} (of worker $i$) (see line 5 of \textit{Boxed Algorithm 2}). These randomly generated points are referred to as \textit{Candidate points} in subsequent discussions. The LMSRBF method in Regis and Shoemaker \cite{Regis2007} introduced the idea of generating candidate points via a Gaussian perturbation of the center point $i$, i.e., $x_{cand} \sim \mathcal{N}(x_{i},\sigma^2I_d)$, where $x_{cand}$ is a candidate point (see Line 9 of procedure 'GenerateCandidates' in \textit{Boxed Algorithm 2}). If a randomly generated candidate point is out of the unit hypercube boundary, it is forced to a corner point. Hence, candidate points are concentrated in a hypersphere around the center point (see Figure \ref{center_visual}(b) for illustration of how candidate points are generated). 

MOPLS employs a similar candidate generation methodology. The standard deviation of the Gaussian perturbation, $\sigma$, is equal to the memory attribute $r_{i}$ ((see Lines 3-4 of procedure 'GenerateCandidates' in \textit{Boxed Algorithm 2}). The memory attribute, $r_{i}$, hence corresponds to the local search neighborhood of the center point, $x_{i}$. Every evaluated point $x_j$ in the set $S_m$ includes an associated memory attribute $r_j$ (see Table \ref{table:nonlin} for definition), which denotes the standard deviation of the Gaussian perturbation for generating candidate points in the neighborhood of $x_j$.  $r_{j}$ is a dynamic variable, which is initialized to be equal to $r_{init}$, and is updated in Step 2.3 of MOPLS (see Figure \ref{fig_framework}) as per the procedure described in \textit{Boxed Algorithm 3} and Section \ref{ch4_tabu}.  


We also implement a variation of the original candidate generation methodology in MOPLS, i.e,  $x_{cand} \sim \mathcal{N}(x_{i}, |\mathcal{N}(\sigma,\sigma/2)|)$ (see lines 5-6 of procedure 'GenerateCandidates' in \textit{Boxed Algorithm 2}). With this variation, candidates are concentrated in a hyperellipse and search concentration varies across the decisions. Such a search methodology might be particularly useful in the presence of varying sensitivities of decision variables. In each iteration of MOPLS, each worker $i$ chooses one of the two candidate generation methodologies with equal probability.
  
 The number of candidate points generated by the worker process (or Worker $i$) is equal to $500*d$, where $d$ denotes the number of decision variables (see lines 7-10 of procedure 'GenerateCandidates' in \textit{Boxed Algorithm 2}). After generation of the candidate points the local surrogate model is used to approximate their objective functions. The \textit{candidate points} approximated by the surrogate model that are non-dominated (as per the surrogate approximation), are identified (see line 6 of \textit{Boxed Algorithm 2}), and a new point is subsequently chosen for evaluation.  

\subsubsection{ Selection of Expensive Evaluation Point}
\label{ch4_rule}
 Let $X^*_{cand}$ denote the set of non-dominated candidate points as per the surrogate approximation. A new point for expensive evaluation is selected from the set $X^*_{cand}$. Selection of a new point for actual evaluation is based on one of two strategies, namely, hypervolume improvement and Max-Min method (see Lines 7-11 of pseudo-code of \textit{Boxed Algorithm 2}). 
 
The hypervolume improvement selection method computes the approximate hypervolume improvement of every RBF approximated non-dominated candidate solution, and chooses the one with the best hypervolume improvement. The hypervolume improvement-based selection method is defined in the Procedure "BestHypervolume" of \textit{Boxed Algorithm 2} (referenced in line 8 of \textit{Boxed Algorithm 2}) and is illustrated in Figure \ref{ch4_hyp_visual}(b). Hypervolume improvement is calculated via a Monte Carlo experiment, in order to reduce computational burden. The approximate hypervolume improvement can quantify the potential value added by a candidate in terms of both convergence and diversity, and hence, selection based on approximate hypervolume is purely based on exploitation of the surrogate model within the local search. Hypervolume improvement based selection is used in various other surrogate and non-surrogate algorithms \cite{Bader}. 
 
 The Max-Min selection methodology is depicted in the Procedure "MaxMinMethod" of \textit{Boxed Algorithm 2} (referenced in line 11). As per the max-min method, an RBF-approximated non-dominated candidate point which is furthest from already evaluated points is selected for function evaluation. While the candidate is chosen from amongst the approximated non-dominated set, i.e, $X^*_{cand}$ (exploitation), it is relatively furthest from the already evaluated points (exploration). Hence, the max-min selection method incorporates both exploration and exploitation. The probability of hypervolume improvement based candidate selection (denoted as $prob_{hv}$ in Algorithm Step 2.2) is set to 0.65, for all experiments. Hence, the value of $prob_{hv}$ is fixed and it is not considered an algorithm parameter.
 
 \subsubsection{Mutation}
\label{ch4_mutate}
The methodology for surrogate-assisted local search is depicted in lines 3-11 of  \textit{Boxed Algorithm 2}. As is evident from the pseudo code depicted in  \textit{Boxed Algorithm 2}, selection of a new point for expensive evaluation is not necessarily based on surrogate-assisted local search. The worker process employs surrogate-assisted local search with a probability, $prob_{cand}$ ,where the value of $prob_{cand}$ is set as 0.9 for all computational experiments (hence it is not considered an algorithm parameter in this study). Each worker $i$ employs mutation as an alternative to surrogate-assisted local search, for selecting a new point for expensive evaluation.

The use of mutation is a critical feature in evolutionary algorithms, ensuring that search remains global and has the potential to move to any point in the decision space. Mutation is also employed in MOPLS (see line 13 of \textit{Boxed Algorithm 2}) to ensure that the search remains global. We employ a Gaussian or a uniform mutation with equal probability. (see Procedure 'mutate' in  \textit{Boxed Algorithm 2}). 

After worker $i$ selects a new point for expensive evaluation (denoted as $x^*_i$), the worker expensively evaluates the selected point (see line 14 of  \textit{Boxed Algorithm 2}). Furthermore, the memory attributes of the newly evaluated point are also initiated (see line 14 of \textit{Boxed Algorithm 2}) and the newly evaluated multi-attribute point (denoted as $z^*_i$ in \textit{Boxed Algorithm 2} and Table \ref{table:nonlin}) is sent to the master process. 
 
 \subsection{Tabu List and Memory Archive}
\label{ch4_tabu} 

In Section \ref{ch4_frame}, we defined the ordered set of evaluated points as $S_m = \{z_i \mid z_i = (x_i, y_i, r_i, c_i, c'_i), 1 \le i \le m\}$, where $z_i$ is a multi-attribute element of $S_m$, $x_i$ is the ith decision variable, $y_i$ is the objective vector corresponding to $x_i$, and $r_i$, $c_i$ and $c'_i$ are memory attributes corresponding to $x_i$. $r_i$, $c_i$ and $c'_i$ (for $1 \le i \le m$) are collectively referred as the memory archive of the evaluated set, $S_m$. The memory attribute $r_i$ is generically referred as the "local search radius" and corresponds to the standard deviation of gaussian perturbations that generate the candidate points of Step 2.2 (discussed in Section \ref{sec_candidates}). The memory attribute $c_i$ is generically referred as the "failure count", and is discussed further later in this section. The memory attribute $c'_i$ is generically referred as the "tabu count", and is also discussed further later in this section. The Step 2.3 of the general algorithm framework of MOPLS updates the memory archive, after new points have been selected and evaluated via the parallel local search of Step 2.2. The procedure for updating the memory archive is depicted in \textit{Boxed Algorithm 3}.

\begin{algorithm}
\fontsize{9.2}{7.2}\selectfont
\LinesNumbered
\NoCaptionOfAlgo
\SetAlgoSkip{smallskip}
\SetAlCapSkip{0ex}
\SetKwData{Left}{left}
\SetKwData{This}{this}
\SetKwData{Up}{up}
\SetKwFunction{HV}{HypervolumeContributions}
\SetKwFunction{FindNonDominated}{\text{FindNonDominated}}
\SetKwFunction{Sort}{SortFront}
\SetKwInOut{Input}{Input}
\SetKwInOut{Output}{Output}
\BlankLine
\Input{1) $S_{m}$ - Ordered set of points evaluated before current algorithm generation.\\  2) $I_{cen}$ - Indices of the \textit{centers} chosen for parallel local search that correspond to the \\ ordered archive of evaluated points, $S_m$. \\3) $S_{new}$ - New points selected and evaluated \\ in \textit{Boxed Algorithm 2}. \\4) $S_{tabu}$ - Set of \textit{tabu} evaluated points. 
\\ 5) $P_{m}$ - Subset of {\it{non-dominated}} in $S_{m}$. }
\BlankLine
\Output{1) $S_{m}$ - Ordered set of points evaluated before current algorithm generation, with \\ updated memory attributes.\\ 2) $S_{tabu}$ - Updated list / set of \textit{tabu} evaluated points. $S_{tabu} \subset S_{m}$. }
\Begin{
\For{$k = 1$ to $|S_{new}|$ {\bf and} $(x^*_k, y^*_k, r^*_k, c^*_k, c'^{*}_k) \in S_{new}$ {\bf and} $j_k \in I_{cen}$}{
$Y_m = \{y_i \mid (x_i, y_i, r_i, c_i, c^{'}_i) \! \in \! P_m, 1 \! \le \! i \! \le \! |P_m| \}$ \;
$HI_k =  \bigg[ H_V(Y_m \cup y^*_k) - H_V(Y_m) \bigg]$\;
\If{$HI_k == 0$}{
$i = j_k$\;
$z_i = (x_i, y_i, r_i, c_i, c^{'}_i) \in S_m$, i.e, $x_i$ is the center point (step 2.1) around which  local search (Step 2.2) was performed to obtain the new evaluated point $x^*_k$ \;
$r_i = \frac{r_i}{2}$ \;
$c_i = c_i + 1$ \;
}
} 

\For{$i = 1$ to $|S_{m}|$ {\bf and} $(x_i, y_i, r_i, c_i, c^{'}_i) \in S_{m}$}{
Let $z_i = (x_i, y_i, r_i, c_i, c^{'}_i)$ \;
\If{$c^{'}_i > 0$}{
$c^{'}_i = c^{'}_i - 1$ \;
\If{$c^{'}_i == 0$}{
$S_{tabu} = \{S_{tabu}\} \backslash \{z_i\}$\;
}
}
\ElseIf{$c_i > c_{thresh}$}{
$c^{'}_i = c_{tenure}$\;
$r_i = r_{init}$\;
$c_i = 0$\;
$S_{tabu} = \{S_{tabu}\} \cup \{z_i\}$\;
}
} 

}
\caption{\textbf{Boxed Algorithm 3: Update Memory Archive}}
\end{algorithm}

The process of updating the memory archive is divided into two phases. In the first phase (depicted by lines 2-9 of \textit{Boxed Algorithm 3}) we traverse through the list of center points that were chosen for local search in Step 2.1 (see Figure \ref{fig_framework}) of current MOPLS iteration, and assess the performance of local search around each center point. Performance of a \textit{center point} is assessed by computing the hypervolume improvement registered by the new point generated from local search around the center point (see line 4 of \textit{Boxed Algorithm 3}). If a hypervolume improvement is not registered, the "local search radius" ($r_i$) is reduced by half, and the "failure count" ($c_i$) is incremented by one. Hence, it can be said that the "failure count", $c_i$, archives the number of times a local search around $x_i$ did not contribute to an improvement in the non-dominated solution set.  When the "failure count" of a point exceeds the number, $c_{thresh}$, the center is added to the Tabu list, $S_{tabu}$ (discussed further in the following paragraph).

As mentioned earlier, the memory attribute $c'_i$ is referred as the "tabu count" and corresponds to the number of algorithm iterations remaining for which a point $x_i$ will remain in the tabu list, $S_{tabu}$. A non-zero value of "tabu count", thus, implicitly means that the point is in the Tabu list. In the second phase of the memory archive update procedure (depicted in lines 10-20 of \textit{Boxed Algorithm 3}) we traverse through all evaluated points (not including the points evaluated in the current algorithm generation) to 1) update the tabu list, and 2) update the tabu count. For all points which are in the tabu list the "tabu count" is reduced by one. If the "tabu count" for a point in the tabu list is reduced to zero, the point is removed from the tabu list. For points which are not in the tabu list, we check if the "failure count" exceeds the number, $c_{thresh}$. If this is the case, the point is added to the tabu list ($S_{tabu}$), the "tabu count" is set equal to the number, $c_{tenure}$, the "failure count" is reset to zero, and the "local search radius" is set equal to the initial search radius, $r_{init}$ (Note that $r_{init}$ is an input parameter for MOPLS). It can be deduced from the above discussion that $c_{tenure}$ is the tabu tenure for a point that is added to the tabu list.   

The self-adaptive failure count method (inspired by Regis and Shoemaker \cite{Regis2007}) ensures that the algorithm does not get stuck in searching around locally optimal solutions, and tends to promote \textit{exploration} within the search mechanics. Since our parallel search methodology revolves around effective selection of \textit{centers}, exploration within center selection could be vital to maintain a global search. However, we would not like to discard a point as a potential center forever. Hence, after a fixed number of algorithm iterations, $c_{tenure}$, a tabu center is removed from the \textit{tabu} list. In all our experiments, $c_{thresh}$ and $c_{tenure}$ are set to 3 and 5, respectively.
 

 \section{Computational Experiments}
Assessing the performance of a Multi-Objective Optimization (MOO) algorithm is considerably more challenging than assessing the performance of a single-objective optimization algorithm both because there are more factors to evaluate for a multi-objective problem and because each trial is more computationally expensive. Coello et al. \cite{CoelloCoello2007} recommend that the quality of a MOO algorithm be assessed via two major factors: efficiency and effectiveness. Efficiency signifies the speed with which the algorithm obtains good quality solutions, and effectiveness measures the quality of solutions, addressed in terms of both convergence and diversity. We assess the efficiency and effectiveness of MOPLS by comparing it against some benchmark algorithms with application to various test problems and real world applications. The remainder of this section summarizes the experimental setup employed to assess the performance of MOPLS.


 
 \subsection{Alternate Algorithms}
 The performance of MOPLS is compared to GOMORS \cite{Akhtar2015}, ParEGO \cite{IMSExample:Knowles2006}, Borg \cite{Hadka2013}, NSGA-II \cite{IMSExample:Deba}, NSGA-III \cite{Deb2014A} and MOEA/D \cite{IMSExample:Zhang2007}. GOMORS is a Radial Basis Function based iterative surrogate optimization algorithm \cite{Akhtar2015}. ParEGO \cite{IMSExample:Knowles2006} is a popular kriging based efficient MOO algorithm. Borg is a steady-state multi-method evolutionary algorithm built on the structure of $\epsilon$-MOEA, that employ various search mechanics for efficient evolutionary optimization of multi-objective problems. 
 
 NSGA-II, NSGA-III and MOEA/D are population-based evolutionary algorithms that evolve in population batches, i.e., an entire new population is evaluated in each algorithm iteration. Hence, these algorithms can easily be parallelized in a batch-parallel synchronous framework.
 
 
Four different versions of MOPLS are employed in the analysis; MOPLS-4, MOPLS-16, MOPLS-32 and MOPLS-64, where the $N$ in MOPLS-N refers to the number of simultaneous centers. Hence, MOPLS-16 employs 16 simultaneous centers (i.e, 16 synchronous parallel processes within each algorithm iteration), and local surrogate-assisted search is performed around each center. MOPLS, in a synchronous master-slave parallel framework, could allow N simultaneous local searches to be executed if N cores are available.  
 
 \subsection{Test Problems}
 \label{ch4_testproblemdescription}
We have chosen eleven unconstrained test problems for algorithm performance analysis. All of the test problems chosen pose different challenges to the search optimization process. Hence, if performance of an algorithm is consistently good across all problems, one can be more confident of the algorithm's capabilities.

The first five test problems, ZDT1, ZDT2, ZDT3, ZDT4 and ZDT6 are part of ZDT test suite  \cite{IMSExample:Deb1999}. ZDT3 has a disconnected Pareto front, ZDT4 is highly multi-modal and ZDT6 has a low density of solutions, around the Pareto front. Additionally, the distribution of solutions on the ZDT6 Pareto front is non-uniform, and the Pareto front is non-convex. 

The next six test problems used in our study (LZF1-LZF6) were proposed by Li et al. \cite{Li2009}, and highlighted by \cite{Zhang2009}, as potentially hard MOO problems, where the Pareto-optimal sets are complicated and difficult to find. 

All test problems have two objectives and the number of decision variables can vary between 2 and 30.  We test and compare performance of MOPLS on the test problems with 8, 16 and 24 decision variables. This is done to assess the differences in performance of MOPLS with other algorithms on low (less than 10 decision variables) and high (greater than 10 decision variables) dimensional problems. 

\subsection{Watershed Model Calibration Problems}
\label{sec_watershed_definition}
In an effort to analyze performance on real world problems, we employ MOPLS and alternative algorithms for Multi Objective calibration of the SWAT Townbrook watershed model developed by Tolson and Shoemaker \cite{Tolson2007}. Townbrook is a sub-watershed of the Cannonsville basin in upstate New York, and has a drainage area of 37 km\textsuperscript{2}. SWAT \cite{SWAT2005a} is the most widely used distributed watershed model in the world. It is used by government and consultants for making practical water management decisions as well as by hydrologic researchers.

A critical modeling component of SWAT Townbrook is the hydrologic simulation of flow through the sub-watershed in response to rainfall. We attempt to calibrate 15 parameters of the SWAT Townbrook watershed model by formulating the flow calibration as a Multi-Objective global optimization problem. Two alternative formulations are tested, i.e., a bi-objective calibration formulation called TBROOK1 (see Table \ref{table:chapter4_formulations}) and a three-objective formulation called TBROOK2 (see Table \ref{table:chapter4_formulations}). 10 years of real data (from United States Geological Survey (USGS) Station 01421618) is used in both calibration formulation case studies (see Table \ref{table:chapter4_formulations}, and the calibration objectives for each formulation are different metrics that represent error between simulation output and observed data.  

Simulation times for watershed models can vary between order of seconds to order of minutes, depending on the modeling time horizon and the watershed area. Townbrook is a relatively small watershed that has a simulation time of around 10 seconds for a 10 year simulation time horizon.  For larger watersheds simulation times are in order of minutes.


\begin{table}
\caption{The flow calibration formulations employed in comparative algorithm analysis} 
\centering 
\begin{tabular}{| c | c | c | c |} 
\hline 
\bf{Problem} & \bf{Decision} & \multirow{2}{*}{\bf{Objectives}} & \bf{Simulation} \\ 
\bf{Name} & \bf{Variables}  &  & \bf{Time/Eval(s)} \\ [0.3ex] 
\hline 
\textit{TBROOK1} & 15 & 2 & 10\\ 
\textit{TBROOK2} & 15 & 3 & 10\\
 [1ex] 
\hline 
\end{tabular}
\label{table:chapter4_formulations} 
\end{table}

\subsection{Algorithm Comparison Setup}
\label{ch4_sec_compmethod}

\subsubsection{General Experimental Setup}

Since all algorithms are stochastic, we have performed multiple trial runs of each algorithm on all test problems and the watershed calibration problems. 10 trials each are performed for all algorithms on the test problems and the watershed problems. Efficiency of all algorithms is compared in serial ,i.e., w.r.t number of expensive evaluations, and in parallel, i.e., w.r.t wall clock time (see Section \ref{wall_setup} for definition). The parallel comparison is important since MOPLS is designed to be efficient for computationally expensive problems both as a serial and a parallel algorithm, including cases with a large number of processors.

\subsubsection{Evaluation Budget - Serial}
Serial algorithm performance comparison is limited to a maximum of 400 function evaluations for the test problems and 600 function evaluations for the watershed calibration problems.   

\subsubsection{Evaluation Budget - Parallel}
\label{wall_setup}
An advantage of MOPLS is the added efficiency that might be obtained via parallelization. Hence, we also use wall clock time to effectively compare performance of algorithms in a parallel setting. Our definition of wall clock time is defined in the following paragraph and is valid with the assumption that function evaluations are computationally expensive. We have to make this assumption since we compare algorithms on test problems that are not computationally expensive.
 
Lets assume that that computation time for each function evaluation is sufficiently high such that other computational overheads arising, for example, from updating the response surface, or communication between processors are negligible. This is reasonable since expensive functions can take from many minutes to hours for each evaluation. Furthermore, we assume that $N$, i.e, the number of expensive function evaluations being evaluated in one algorithm iteration can be evaluated simultaneously. This means that if $N_{core}$ is the number of cores available for parallel computing, $N_{core} = N$. Given these assumptions, one unit of wall clock time is equivalent to one algorithm iteration.  

We set the maximum wall-clock time to be $60$ units, excluding time for initial sampling and evaluations. The total function evaluations will consequently be equal to $(2d +2) + (W*N)$, where $d$ is number of decision variables and $(2d+2)$ is the number of initial function evaluations. $W$ is the wall-clock time budget (or synchronous parallel algorithm iterations or number of generations for evolutionary algorithms) defined to be equal to 60 in our experiments. $N$ is the number of expensive function evaluations being evaluated in one algorithm iteration. Alternatively, $N$ may also be referred as the batch size or population size.

We compare multiple algorithms running in parallel or serial. ParEGO is a serial algorithm. Hence, $N =1$ for ParEGO. GOMORS is parallel and four expensive evaluations are performed in each algorithm iteration. Hence, $N =4$ for GOMORS. In case of MOPLS-N, we can choose to have many cores. Hence, we choose $N$ = 4, 16 and 64. For the evolutionary algorithms compared in parallel, $N$ = 16 and 64. Note that Borg-MOEA is not compared to other algorithms in the synchronous parallel comparison framework (Borg is only compared in serial), since it is a non-stationary evolutionary algorithm, and hence can only be compared fairly in an asynchronous parallel comparison setting.   




\subsubsection{Algorithm Parameters}

Our algorithm comparison methodology incorporates the use of suitable values for parameters of all algorithms under discussion with the aim of conducting a fair comparison. Hence, default parameter configurations of ParEGO \cite{IMSExample:Knowles2006} and GOMORS \cite{Akhtar2015} are used. 

Borg MOEA has an adaptive population size and hence, does not require tuning of the population. However, the default minimum, initial and maximum population sizes of Borg MOEA are $100$, $100$ and $10,000$ respectively. Since the function evaluation budge, for Borg, in this diagnostic algorithm comparison is limited to 600, these values for the initial, minimum and maximum population sizes are not desirable. Given the limited evaluation budget, we changed the initial, minimum and maximum population size values for Borg MOEA to $16$, $16$ and $64$, respectively.

\begin{table}
\caption{Parameter Settings for MOPLS} 
\centering 
\begin{tabular}{| c | l | r | } 
\hline 
\bf{Name} & \bf{Description} & \bf{Setting} \\ [0.5ex] 
\hline 
 $E_I$ & The number of initial expensive evaluations & $2d+2$  \\  
$E_T$ & The number of total expensive evaluations & 400 or 1000 \\ 
$N$  & The number of \textit{center points} & 4, 8, 16, 32 or 64 \\
$r_{init}$ & Candidate / local search parameter /radius & 0.2  \\
\hline 
\end{tabular}
\label{tab_par_mopls} 
\end{table}

As mentioned in Table \ref{table:params}, there are four parameters in the MOPLS algorithm. MOPLS' parameter configuration is provided in Table \ref{tab_par_mopls}. As indicated in Section \ref{ch4_frame} (Step 1), the number of initial function evaluations is fixed at $(2d + 2)$ as per the recommendations provided by Regis and Shoemaker \cite{Regis2007}, where $d$ is the number of decision variables. This is equal to the number of initial function evaluations of GOMORS and ParEGO.  Value of the candidate search radius, i.e., $r_{init}$, for MOPLS is set to 0.2 for all test instances. This value of the search radius is recommended by Regis and Shoemaker \cite{Regis2007}.  

 
   \begin{figure*}[!ht]
 \centering
 \noindent\includegraphics[width=43pc]{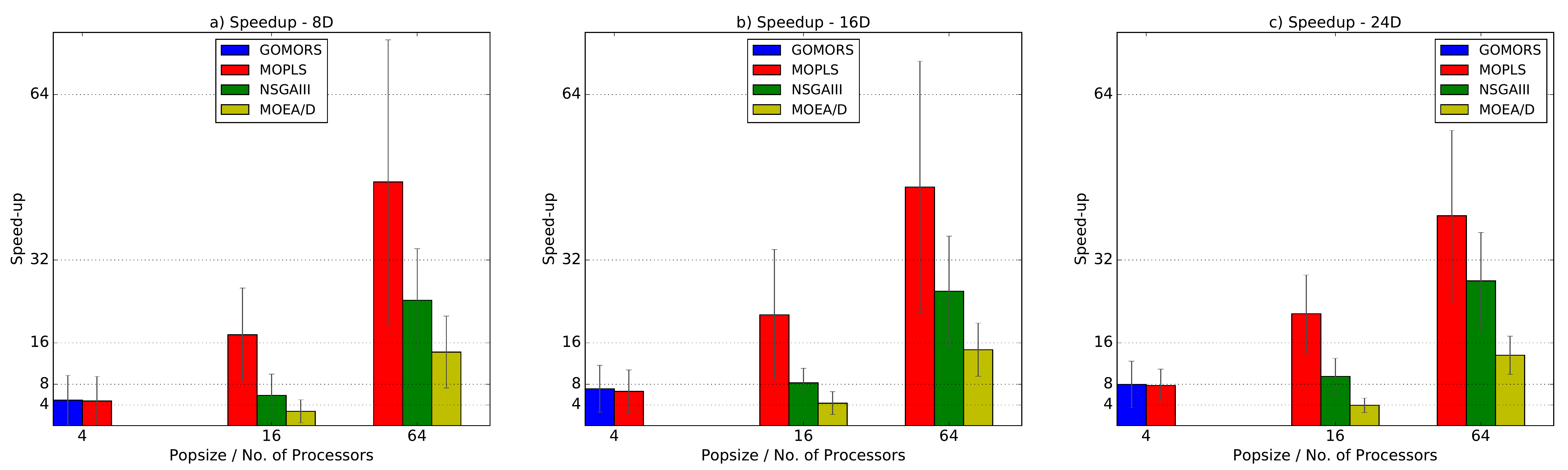}
 \caption{\textit{\textbf{Speed-up Plots}}: Speed-up (Eq. \ref{speedup_definition}) of algorithms in terms of the average (over multiple trials) number of parallel iterations (generations) taken by each algorithm to reach the average (over multiple trials) hypervolume precision obtained by ParEGO (serial) after 400 function evaluations. Speed-ups are further averaged over all eleven test problems (see Section \ref{ch4_testproblemdescription}) and reported in the bar-charts above, for each algorithm and different number of processors (4-64). Each sub-figure corresponds to the a fixed number of problem dimensions for all test problems (8,16 and 24). Higher speed-ups values are desirable. Speed-up variations over test problems are indicated by error bars.}
 \label{fig_speedup}
 \end{figure*}

\subsubsection{Performance Metric}
\label{sec_hc_definition}

The key performance metrics employed in the analysis is the hypervolume coverage (defined in Equation \ref{hc_definition}). Figure \ref{ch4_hyp_visual} (also see Table \ref{table:nonlin} for definition) gives a visual illustration of the meaning of Hypervolume, denoted as $H_v(P)$. Let $P$ be the set of non-dominated solutions obtained as an approximation to the Pareto front from an algorithm. Moreover, let $P^*$ be the set of ideal  solution(s) of the MOO problem being solved. The "ideal solution(s)" of a MOO problem is the Pareto front, if known (the Pareto front is known for the test problems). If the Pareto front is not known (as is the case for the watershed problems), $P^*$ is assumed to be the "estimated" best non-dominated front obtained from all algorithms across all trials. The "reference point" depicted in Figure \ref{ch4_hyp_visual} is the worst attainable solution of the optimization problem. Since, the worst attainable values are not known in the problems discussed in this study, the "reference point" vector is estimated as the worst values of all objectives obtained in our computer experiments. Hence, the total feasible objective space is the volume bounded by the reference point and ideal solution(s). 


The Hypervolume coverage is defined as follows:

\begin{equation}
 H_c = \frac{H_v(P) - H_v(P_{init})}{H_v(P^*) - H_v(P_{init})}
 \label{hc_definition}
 \end{equation}

 As per the above equation, Hypervolume Coverage, i.e., $H_c$, is the fraction of total feasible objective space (after subtracting the objective space dominated by the initial solutions) dominated by estimate of the Pareto front obtained by an algorithm. A higher value of $H_c$ indicates a better solution and the ideal value is $1$. The Hypervolume coverage incorporates both convergence to the ideal solution(s) as well as diversity of solutions within a single metric and also discounts any hypervolume domination achieved by the initial solutions.

%

\section{Results and Discussion}
\subsection{Parallel Speed-up Analysis - Test Problems}
\label{sec_speedup}

Since MOPLS is a surrogate algorithm designed to be efficient in a synchronous parallel setting, an important metric for analyzing parallel performance of MOPLS is \textit{speed-up}. We formally define speed-up in this analysis as follows:

\begin{equation}
 S(a,p,N,\alpha_{T(b,p,1)}) = \frac{T(b,p,1)}{T(a,p,N)}
\label{speedup_definition}
\end{equation} 

Let $T(a,p,N)$ be the `time' required by algorithm $a$ to 'solve' (i.e., to achieve a target Hypervolume coverage, $\alpha$) the problem $p$ with $N$ processors. In our analysis, 'time' is defined as number of function evaluations for serial algorithms (i.e., $n=1$) and wall clock time (as defined in Section \ref{wall_setup}) for parallel algorithms. In the above equation, $\alpha = \alpha_{T(b,p,1)}$ is the prescribed target Hypervolume Coverage ($H_c$), that is defined as the average Hypervolume Coverage achieved by the baseline serial algorithm $b$, for problem $p$, after a fixed $T(b,p,1) = T$ function evaluations. Consequently, speed-up of an algorithm $a$, for problem $p$, for the target $H_c = \alpha_{T(b,p,1)}$, is the serial time (i.e., $T(b,p,1)$) required by the baseline algorithm to reach $\alpha_{T(b,p,1)}$ hypervolume coverage, divided by the number of wall clock time units, i.e., $T(a,p,N)$,  required by algorithm $a$ to reach $\alpha_{T(b,p,1)}$ hypervolume coverage (for problem $p$), where $N$ is the population size (or number of simultaneous evaluations per iteration) for algorithm $a$. 

We choose ParEGO as the baseline algorithm $b$, for our speed-up analysis, since it is a well-known and efficient serial surrogate MOO algorithm. Furthermore, since we are interested in analyzing algorithm performance on a limited evaluation budget, the target hypervolume coverage, $\alpha_{T(b,p,1)}$, is defined as the average coverage achieved (over multiple trials) by ParEGO in 400 function evaluations (i.e., $T(b,p,1) = 400$).

Given Equation \ref{speedup_definition} and the definition of $\alpha_{T(b,p,1)}$ provided in the preceding paragraph, speed-ups for all combinations of algorithms (denoted by $a$) and number of simultaneous evaluations (or population size, denoted by $N$), are analyzed, for all test problems (denoted by $p$) introduced in Section \ref{ch4_testproblemdescription}. Speed-up values for each algorithm (with $N$ evaluations per parallel iteration) are averaged over multiple trials for each test problem to obtain an average speed-up value for an algorithm on a particular problem. We refer this speed-up value as $\hat{S}(a,p,N)$ in subsequent discussions.

Figure \ref{fig_speedup} provides a summary of speed-ups of all synchronous parallel algorithms (applied to the eleven test problems mentioned in Section \ref{ch4_testproblemdescription}), with ParEGO as the baseline algorithm and average $H_c$ obtained by ParEGO after 400 function evaluations as the target precision.

Since we tested all algorithms on multiple test problems (see Section \ref{ch4_testproblemdescription}) with 8 ,16 and 24 decision variables, the speed-up results are grouped for a fixed number of decision variables. Consequently, each sub-figure reports speed-ups of algorithms for a fixed number of decision variables (8, 16 and 24). Furthermore, since algorithms are applied to multiple test problems for a fixed number of decision variables, the bars in each sub-figure of Figure \ref{fig_speedup} report the average speed-ups over a single problem (defined as $\hat{S}(a,p,N)$ in the previous paragraph), averaged over all 11 test problems (for a fixed number of decision variables). The error margins of each bar (depicted by black lines) report the standard deviation of algorithm speed-up (over multiple test problems).   

 \begin{figure*}[!ht]
 \centering
 \noindent\includegraphics[width=43pc]{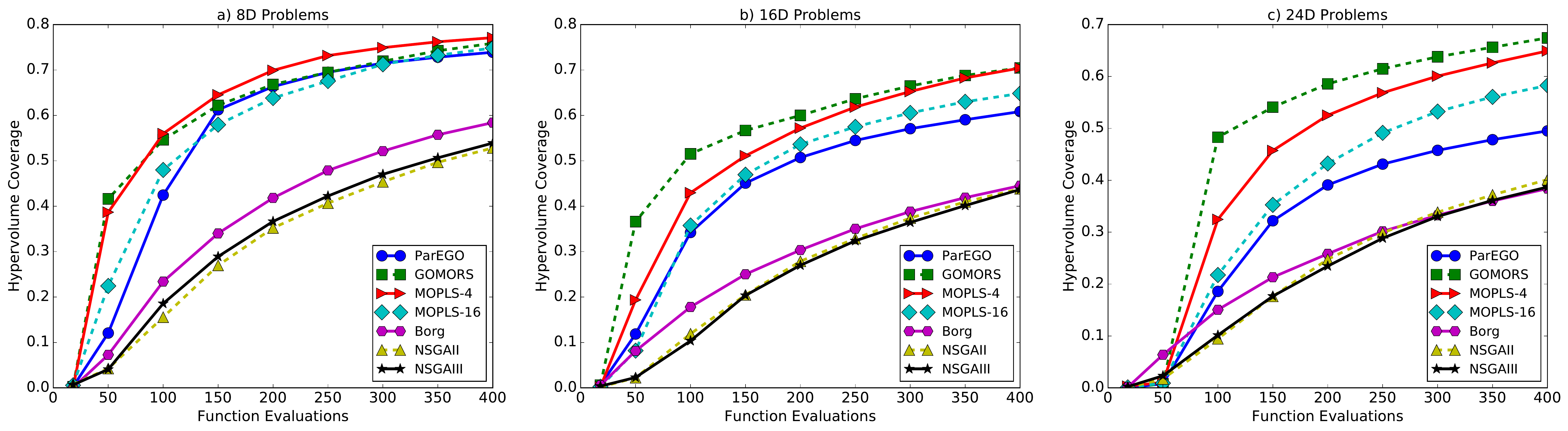}
 \caption{\textit{\textbf{Average Serial Progress Plots}}: Comparison of Hypervolume coverage, $H_c$, progress, for all algorithms, as a function of function evaluations (serial), averaged over all test problems (defined in Section \ref{ch4_testproblemdescription}) with 8, 16 and 24 decision variables. Each subplot corresponds to a Hypervolume coverage progress plot (higher values are better) comparison of all algorithms on all test problems for a fixed number of decision variables (depicted in subplot title).}
 \label{ch4_serial}
 \end{figure*}

\underline{\it{Results for 64 Processors:}} MOPLS is designed to be effective even when the number of processors is relatively large. Hence, we first analyze speed-ups of algorithms with number of simultaneous evaluations (or processors), i.e., $N$, equal to 64 (see Figure \ref{fig_speedup}). The algorithms compared for $N = 64$ processors are MOPLS, NSGA-III and MOEA/D. GOMORS is not included in this comparison, since it is only designed for $N = 4$. Figure \ref{fig_speedup} clearly shows that MOPLS is the best performing algorithm as per average speed-ups for 8, 16 and 24 dimensional test problems, when $N = 64$. Moreover, MOPLS-64 has approximately 1.5-2 times (depicted by error bars) better average speedup than NSGA-III-64 and 3 times better speedup than MOEA/D-64. 

\underline{\it{Results for 16 Processors:}} Figure \ref{fig_speedup} also compares MOPLS against NSGA-III and MOEA/D when $N=16$. The speed-up analysis for $N =16$ shows that MOPLS-16's average speed-up over serial ParEGO is super-linear. This indicates that MOPLS-16 is, on average (over multiple problems and trials), at least 16 times more efficient than ParEGO, in parallel. Furthermore, when number of decision variables increase, the average speed-up of MOPLS improves further and the standard deviation (over multiple test problems) of the speed-up also reduces. MOPLS-16 has the best speed-up amongst all algorithms compared for $N=16$ and its average speed-up is, approximately, more than two times better than NSGA-III-16 and four times better than MOEA/D-16. 

\underline{\it{Results for 4 Processors:}} Figure \ref{fig_speedup} also compares algorithm speed-ups for number of simultaneous evaluations (or processors), i.e., $N$, equal to 4. Only GOMORS and MOPLS have algorithm settings such that  $N = 4$ (since a population size of 4 is not effective for MOEAs). It is evident in Figure \ref{fig_speedup} that average speed-ups (over multiple problems and trials) of both GOMORS-4 and MOPLS-4 with ParEGO as baseline are very good. The average speed-ups for both GOMORS-4 and MOPLS-4 are in-fact either linear or super-linear in all sub-figures, i.e., for all problems sub-groups (8, 16 and 24 decision variable sub-groups). Furthermore, average speed-up numbers of both algorithms are quite similar. GOMORS-4 is slightly better. However the standard deviation error bars for the 24-dimensional test problems ( see Figure \ref{fig_speedup}(c)) show that speed-up variation is less for MOPLS-4 which indicates that MOPLS-4 has a slightly better reliability (over different problems) than GOMORS-4 on the higher dimensional problems. Another interesting finding here is that speed-ups of both GOMORS-4 and MOPLS-4 increase with increase in decision variables (efficiency of both algorithms is $ \ge 100\%$), indicating that these RBF based surrogate algorithms are more effective than the Gaussian Process based ParEGO for higher dimensional problems.

 \begin{figure*}[!ht]
 \noindent\includegraphics[width = 43pc]{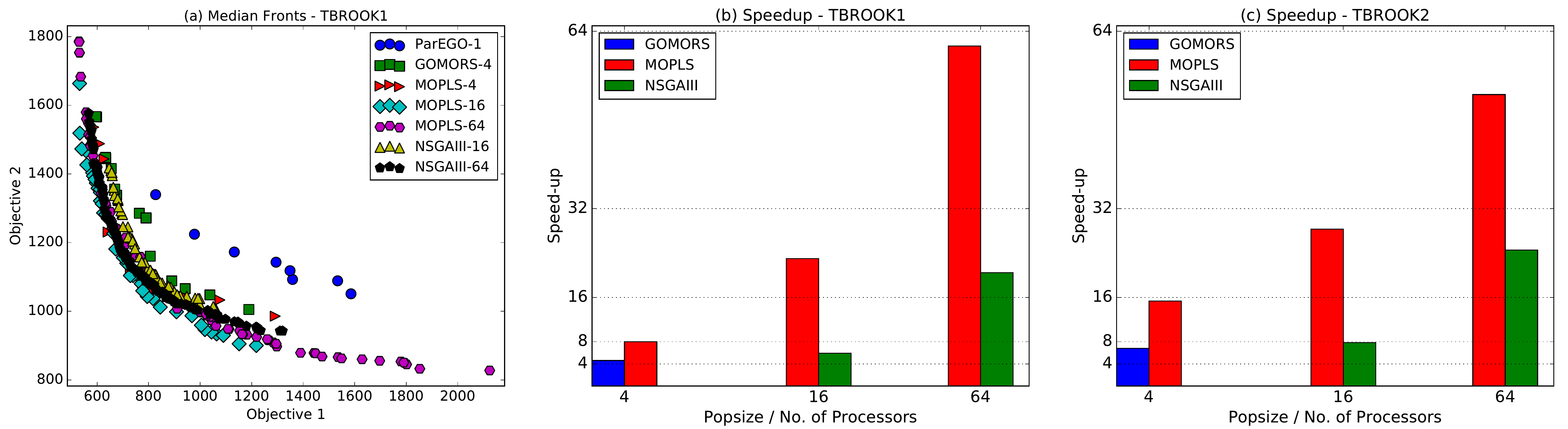}
 \caption{\textit{\textbf{Results for Watershed Problems (see Table \ref{table:chapter4_formulations} for definition)}}:  a) Visualisation of median (as per Hypervolume Coverage) non-dominated fronts from all algorithms (Number of processors used by each algorithm is reported after '-' in the legend.) after 60 wall clock time units (or parallel algorithm iterations) for TBROOK1 bi-objective problem (fronts closer to origin are better), b) Speed-up of algorithms for problem TBROOK1, and c) Speed-up of algorithms for problem TBROOK1. Speed-up (see Eq. 3) is defined in terms of the average (over multiple trials) number of parallel iterations (generations) taken by each algorithm to reach the average (over multiple trials) hypervolume precision obtained by ParEGO (serial) after 400 function evaluations. Speed-ups are reported in the bar-charts above (in subplots b and c), for each algorithm and different number of processors (4-64).  }
 \label{fig_watershed}
 \end{figure*}


The speed-up analysis of Figure \ref{fig_speedup} shows that MOPLS is, overall, the most efficient amongst all algorithms compared for number of simultaneous evaluations ranging from 4 to 64, in terms of parallel speed of convergence to the solution obtained by ParEGO in 400 serial evaluations. Moreover, the analysis indicates that the novel strategy introduced in MOPLS where simultaneous surrogate-assisted candidate searches are performed around multiple (distinct) centers to select and evaluate a batch of new points in each algorithm iteration, is effective and efficient and scales well for up to 64 processors.   

\subsection{Serial Progress Plots - Test Problems}
\label{sec_serial_analysis}
  
This analysis also compares MOPLS against other algorithms in a serial setting, i.e., as a function of number of evaluations. For this purpose, we visualize the average Hypervolume Coverage ($H_c$) of all algorithms as a function of number of evaluations. Such plots are  called progress plots.   

Figure \ref{ch4_serial} plots the 'average' (aggregated over multiple test problems) Hypervolume Coverage ($H_c$) of different algorithms against number of function evaluations. $H_c$ is a normalized metric (Eq. \ref{hc_definition}); hence it is possible to report an aggregated $H_c$ value for an algorithm that is averaged over multiple test problems. Hence, each sub-plot of Figure \ref{ch4_serial} shows aggregated $H_c$ progress plots for different algorithms for a fixed number of decision variables. Note that a higher aggregated $H_c$ value is desirable. Furthermore, the number reported after the algorithm name in the legend represents the population size (or  number of evaluations per iteration) of the algorithm for serial comparison. (In parallel computation comparison the number of processors equals the population size). 


The serial progress comparison results of Figure \ref{ch4_serial} indicate that the RBF-based algorithms, i.e., MOPLS-4, MOPLS-16 and GOMORS are all more efficient than the other algorithms. Moreover, it is evident from Figure \ref{ch4_serial} that performance of evolutionary algorithms, i.e., Borg, NSGA-II and NSGA-III is considerably worst than performance of surrogate algorithms (i.e., ParEGO, GOMORS and MOPLS). Figures \ref{ch4_serial}(b) and \ref{ch4_serial}(c) show that performance of ParEGO becomes significantly worst with increase in number of decision variables. Hence, GOMORS and MOPLS are more suitable than ParEGO (in serial) for problems with more than 10 decision variables. Serial performance of GOMORS is slightly better than that of MOPLS for the 16 and 24 dimensional problems. However, MOPLS can use a large number of processors in parallel while GOMORS is limited to a maximum of 4 processors. Hence MOPLS can achieve much shorter wall clock times (i.e. higher speed-up) for  all algorithm-processor number combinations considered with more than 4 processors (Figure \ref{fig_speedup}).

\subsection{Watershed Calibration Problems}

This section provides an analysis of the effectiveness and efficiency of MOPLS in producing good trade-off solutions within a limited evaluation budget when applied to the Watershed Calibration problems (see Table \ref{table:chapter4_formulations} and Section \ref{sec_watershed_definition} for description of watershed problems)). Our analysis with test problems (Section \ref{sec_speedup}) indicates that increasing the number of local search centers (and with it, the number of processors) in MOPLS results in improved algorithm performance when results are analyzed in wall clock time (especially for MOPLS-16). An increase in the number of centers also results in an increase in the number simultaneous local searches in different regions of interest, and subsequently increases the potential global nature of the search. We wish to see if the same trend is true for the real world watershed calibration problems. 

 Figure \ref{fig_watershed} provides a comparative analysis of all algorithms in wall clock time, with application to the two watershed calibration problems, i.e., TBROOK1 and TBROOK2 (see Table \ref{table:chapter4_formulations}). Figure \ref{fig_watershed}(a) visually compares the median (over multiple trials, according to the hypervolume coverage metric) non-dominated front obtained by each algorithm with application to the bi-objective TBROOK1 problem, after 60 wall clock units. Since both objectives are to be minimized, fronts closer to the origin are desirable. Figure \ref{fig_watershed}(a) shows that the median front of MOPLS-64 is the best. Moreover, the median front of MOPLS-16 dominates the median front of NSGA-III-64, indicating that MOPLS-16 obtains a better trade-off with only $\frac{1}{4}$ the number of processors as NSGA-III-64.  

Figures \ref{fig_watershed}(b) and \ref{fig_watershed}(c) report speed-ups (see Eq. \ref{speedup_definition}) of MOPLS, GOMORS and NSGAIII for watershed problems TBROOK1 and TBROOK2, respectively, with average $H_c$ obtained by ParEGO after 400 function evaluations as the target precision. It is evident here that MOPLS is the most efficient algorithm in parallel (with 4, 16 and 64 processors), for the watershed calibration problems. Speed-ups of  MOPLS-4 and MOPLS-16 are super-linear for both TBROOK1 and TBROOK2. Moreover, both MOPLS-4 and MOPLS-16 have better speed-ups than NSGA-III-16, and NSGA-III-64, respectively, i.e., with MOPLS having only $\frac{1}{4}$ the number of processors as the corresponding NSGA-III parallel case. Speed-up of MOPLS-64 is at least two times better than the speed-up of NSGA-III-64 for both TBROOK1 and TBROOK2.    
  
  
\section{Conclusion}
Multi-Objective Optimization typically requires many more objective function evaluations to obtain accurate Pareto Fronts than required to obtain an accurate solution to a single objective problem. Most MOO algorithms currently widely used  are designed  for inexpensive objective functions.  These algorithms  are  not necessarily feasible or useful for MOO of expensive objective functions requiring solution with a relatively small number of objective function evaluations.

MOPLS-N is an algorithm developed to address the problem of MOO for expensive objective functions by using a combination of  a) surrogate approximation of the objective functions and b) an efficient strategy for selecting in each iteration multiple ($N$) points $\{x_i \!\! \mid \!\! 1\!\! \le \!\! i \!\! \le \!\! N\}$ in the domain space for which the expensive multi-objective function vector $F(x_i)$ is simultaneously computed by $N$ processors.

It is difficult to determine in each iteration (generation) how to best  select $N$  points. MOPLS is designed as a generational population-based algorithms and uses a number of steps for selection of $N$ new points for evaluation in each iteration. These steps include 1) a novel mechanism for selection of a population of $N$ center points for local surrogate search that uses non-dominated sorting coupled with an adaptive radius constraint and a tabu list, and  2) independent and simultaneous surrogate-assisted candidate searches around the $N$ center points to choose and evaluate $N$ new points. (One new point is chosen for evaluation around each center.)


 With regard to parallelism, the master process within the synchronous parallel framework selects the population of center points and maintains the tabu list (which requires little time)  while the worker processes perform surrogate-assisted local search, evaluate new points $F(x_i)$(expensive) and send the information back to the master process.

The results presented here demonstrate that MOPLS is the best parallel algorithm for greater than 4 parallel processors when evaluation numbers  are restricted (because of expensive objective functions). The two RBF surrogate algorithms MOPL-N (with $N=4$) and GOMORS also outperform all of the other algorithms in serial (Figure \ref{ch4_serial}) and with 4 parallel processors.  Results in parallel (analyzed in terms of wall clock time) show that MOPLS-N's performance is, on average, superior to all other algorithms with 16 and 64 evaluations per iteration (averaged over multiple problems).

These serial and/or  parallel  results involve comparisons to well known MOEAs like NSGA-II, NSGA-III, and Borg, which are non-surrogate algorithms that are not designed for applications with a limited number of objective evaluations. So the comparisons here do not undermine the valuable contributions of these earlier algorithms for cases where the objective function evaluation budget is large. Serial ParEGO does use a surrogate and hence is designed for expensive functions and its serial performance is better than the non-surrogate methods with a limited number of evaluations. However, ParEGO has lower efficiency than the other surrogate algorithms MOPLS-N and  GOMORS-4 (Figure \ref{ch4_serial}), especially for higher dimensional problems (16 and 24). 

Application of MOPLS to a realistic and complex watershed flow calibration using the widely used SWAT hydrologic simulation model and a decade of water flow data for a watershed supplying New York City drinking water shows even better results. MOPLS is superior to all other algorithms (see Figure \ref{fig_watershed}). 

In conclusion MOPLS-N is a valuable contribution to the suite of algorithms for MOO  when needed for global optimization of multimodal expensive objective functions. MOPLS is  especially valuable because it can be used efficiently with a relatively large number of parallel processors.  Researchers have access to this algorithm since the software for MOPLS-N (with synchronous parallel implementation) is available in the open source software toolbox pySOT \cite{Pysot2015}. Further innovations can be introduced in MOPLS via pySOT. For example, other meta modeling frameworks \cite{Deb2018} for surrogate fitting can be explored for many objective optimization.

\section*{Acknowledgment}
The authors started this research at Cornell University  and completed it at National University of Singapore (NUS). At Cornell this research was supported by Dr. Akhtar's Fulbright-HEC Pakistan fellowship and by an NSF grant (CISE 1116298) to Prof. Shoemaker. The support at NUS was from the Singapore National Research Foundation, Prime Minister's Office, Singapore under its Campus for Research Excellence and Technological Enterprise (CREATE) programme (E2S2-CREATE project CS-B) and Prof. Shoemaker's NUS startup grant. 


\ifCLASSOPTIONcaptionsoff
  \newpage
\fi



%
%
%

\bibliographystyle{IEEEtran}
\bibliography{IEEEabrv,mopls_ieee}

%

\begin{IEEEbiography}
[{\includegraphics[width=1in,height=1.25in,clip,keepaspectratio]{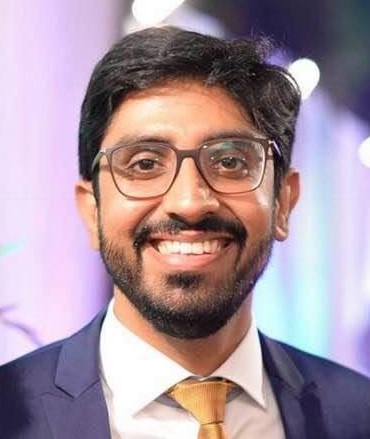}}]{Taimoor Akhtar} is a Research Fellow at NUS' Environmental Research Institute (NERI) in Singapore, and is working with Prof. Christine Shoemaker. Taimoor was a Fulbright Scholar and holds a Ph.D. degree (2015) in Civil and Environmental Engineering from Cornell University, and an M.Eng. degree (2008) in Operations Research and Information Engineering from Cornell University. His research interests are in the application of Multi-Objective Optimization and machine learning to water systems analysis and lake water quality management.
\end{IEEEbiography}

\begin{IEEEbiography}[{\includegraphics[width=1in,height=1.25in,clip,keepaspectratio]{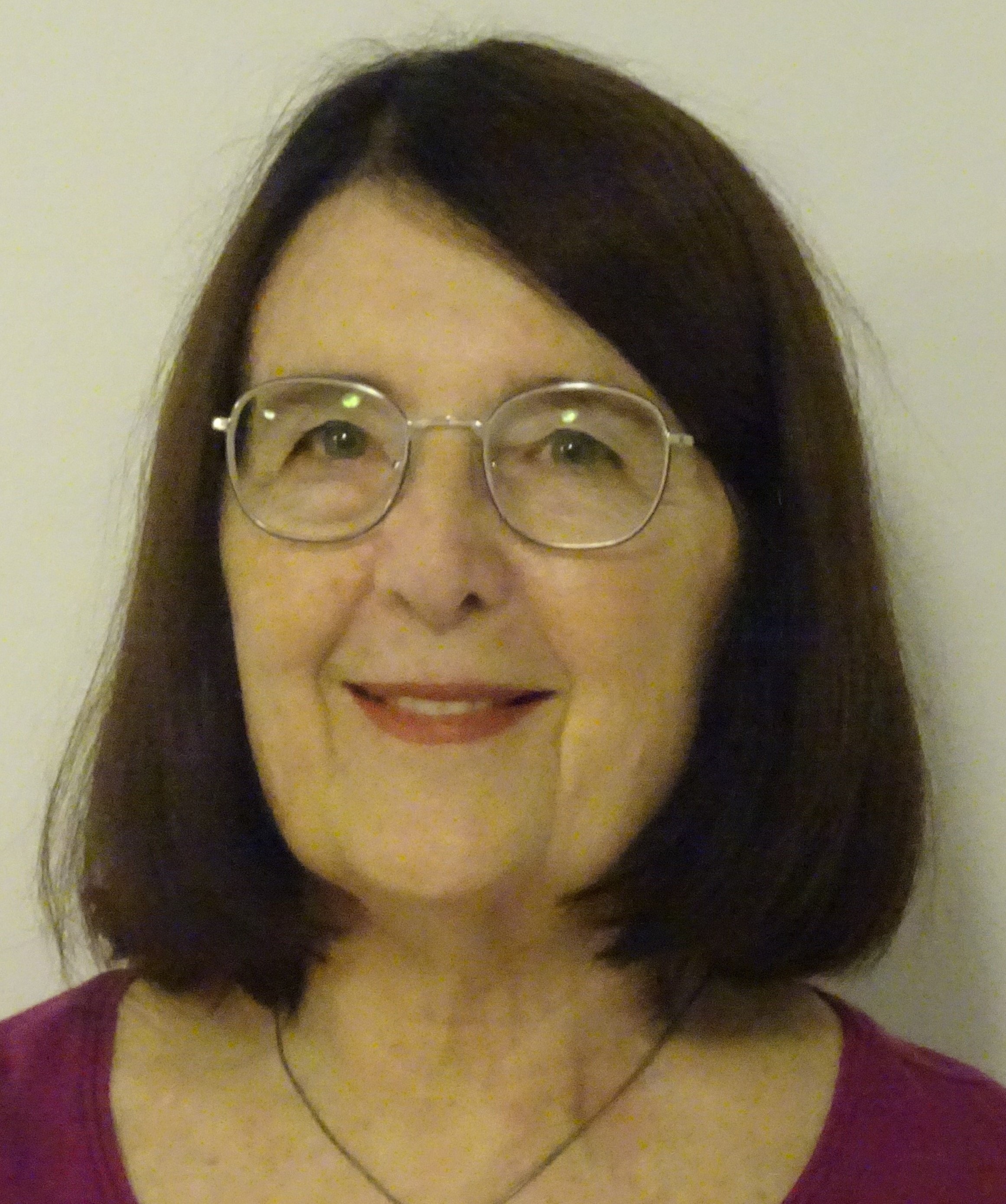}}]{Christine A. Shoemaker}, Ph.D. in mathematics, focuses on global surrogate optimization, stochastic dynamic programming and water resources applications. She is a member of U.S. National Academy of Engineering,  Fellow in SIAM, INFORMS, AGU, and ASCE and Emerita Ripley Professor from Cornell University. She is currently a Distinguished Professor at National University of Singapore.
\end{IEEEbiography}





\end{document}